%% file: main.tex
\documentclass[]{fairmeta}

\usepackage{amsmath,amssymb,amsthm}
\usepackage{graphicx}
\usepackage{float}          
\usepackage{booktabs}
\usepackage{colortbl}
\usepackage{tabularx}
\usepackage{multirow}
\usepackage{subcaption}
\usepackage{xspace}
\usepackage{tikz}
\usetikzlibrary{positioning,shapes.geometric,arrows.meta,calc,patterns,decorations.pathreplacing}

\titleformat*{\paragraph}{\bfseries}

\usepackage[numbers,sort&compress]{natbib}
\usepackage[colorlinks=true,breaklinks=true,
            linkcolor=metablue,citecolor=metablue,urlcolor=metablue]{hyperref}
\ifdefined\pdfsuppresswarningpagegroup\pdfsuppresswarningpagegroup=1\fi

\newcommand{\model}{\texorpdfstring{\mbox{$\mathcal{N}_0$-VTLA}}{N0-VTLA}\xspace}
\newcommand{\modelfull}{\texorpdfstring{\mbox{$\mathcal{N}_0$-VTLA}}{N0-VTLA}\xspace}

\newcommand{\modelbf}{\texorpdfstring{\mbox{$\boldsymbol{\mathcal{N}_0}$\textbf{-VTLA}}}{N0-VTLA}\xspace}
\newcommand{\chizero}{$\chi_0$\xspace}   
\newcommand{\pinought}{\texorpdfstring{$\pi_0$}{pi-0}}
\newcommand{\pizerofive}{\texorpdfstring{$\pi_{0.5}$}{pi-0.5}}
\newcommand{\stageone}{Stage~1\xspace}
\newcommand{\stagetwo}{Stage~2\xspace}
\newcommand{\stagethree}{Stage~3\xspace}

\newif\ifdrafttodos \drafttodostrue
\definecolor{todoorange}{HTML}{C05600}

\newif\ifreviewnotes \reviewnotestrue
\definecolor{revjiongwei}{HTML}{C0392B}     
\definecolor{revfanyt}{HTML}{0E7C7B}      
\definecolor{revlongjie}{HTML}{7D3C98}
\definecolor{revfour}{HTML}{B7950B}       
\definecolor{revfive}{HTML}{C2185B}       

\newcommand{\alter}{\mbox{ALTER}\xspace}

\newtcolorbox{keybox}[1][]{colback=metabg,colframe=metablue,
  boxrule=0.6pt,arc=6pt,left=6pt,right=6pt,top=6pt,bottom=6pt,#1}

\definecolor{metablue}{HTML}{26527A}

\definecolor{vlgreen}{HTML}{CFE1C8}
\definecolor{tactileblue}{HTML}{9ECBD8}
\definecolor{modulepink}{HTML}{FFECE9}
\definecolor{actioncream}{HTML}{F6F1E5}
\definecolor{latentpurple}{HTML}{DFA3D6}
\definecolor{frozengray}{HTML}{D9D9D9}
\definecolor{trainorange}{HTML}{EB6834}
\definecolor{rejectred}{HTML}{D03B3B}
\definecolor{xyzcolor}{HTML}{F6F1E5}
\definecolor{rot6dcolor}{HTML}{E7D6AE}
\definecolor{gripcolor}{HTML}{CDA86B}

\ifdefined\newfontfamily
  \newfontfamily\rlchartfont[
    Path=fonts/,
    Extension=.ttf,
    UprightFont=*-Regular,
    BoldFont=*-SemiBold
  ]{RobotoMono}
\else
  \newcommand{\rlchartfont}{\ttfamily}
\fi
\definecolor{rlpink}{HTML}{FFECE9}
\definecolor{rlcream}{HTML}{F6F1E5}
\definecolor{rlblue}{HTML}{9CCBD7}
\definecolor{rlgreen}{HTML}{CFE1C8}
\definecolor{rlrose}{HTML}{DFA2D6}
\definecolor{rlpurple}{HTML}{CEBAF6}
\definecolor{rlorange}{HTML}{F8CE93}
\definecolor{rlyellow}{HTML}{FBE5AE}

\logo{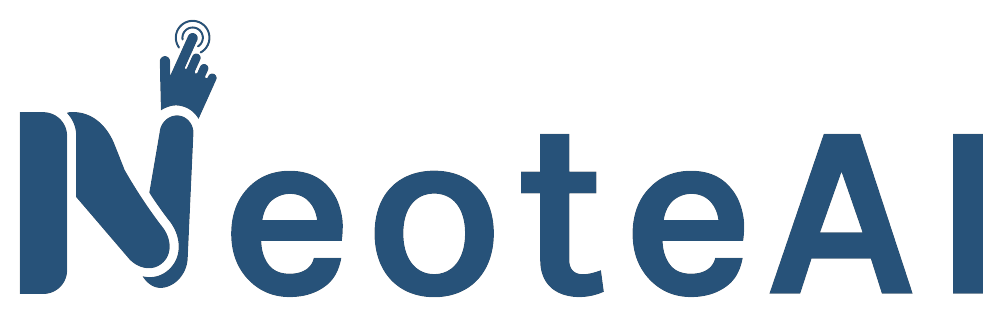}   
\logowidth{2.8cm}         

\title{\modelfull: Scaling Vision--Tactile--Language--\texorpdfstring{\\}{}Action Model with Latent Tactile Tokens}

\author{\large NeoteAI Team \& Fudan TEAI Team}

\date{July 25, 2026}
\metadata[Code]{\url{https://github.com/neoteai/N0-VTLA}}
\metadata[Website]{\url{https://research.neoteai.com/n0-vtla/}}

\begin{document}

\abstract{\input{sections/abstract.tex}}
\logo{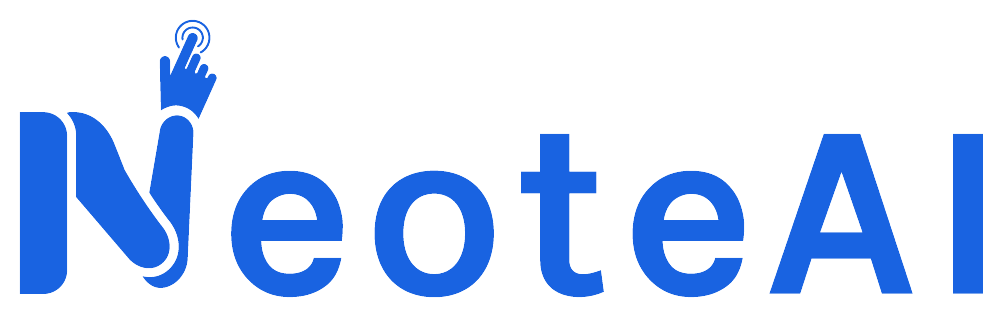}   
\logowidth{2.8cm}         
\toplogo{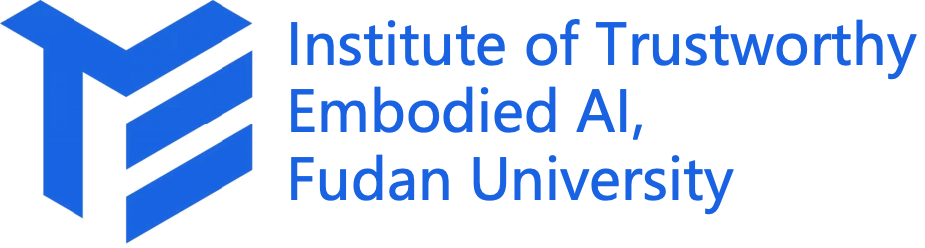}
\toplogowidth{2.8cm} 
\maketitle

\input{sections/intro.tex}
\input{sections/model.tex}
\input{sections/data.tex}

\input{sections/training.tex}

\input{sections/experiments.tex}
\input{sections/findings.tex}

\providecommand{\relatedworkfile}{sections/related.tex}
\input{\relatedworkfile}

\input{sections/conclusion.tex}

\input{sections/limitations.tex}
\input{sections/contributors.tex}

\bibliography{references}

\appendix
\vspace{1em}
\input{sections/appendix.tex}

\end{document}

%% file: sections/abstract.tex
We present \modelfull{}, a vision--tactile--language--action (VTLA)
foundation model capable of (1) fine-grained contact-rich manipulation
with tactile perception and tactile-feedback control, and (2) offline policy improvement from stored deployment data.
Stepping towards current visual-based backbones, we propose an overall training recipe for tactile integration, 
consisting of visuo-tactile pre-training, staged tactile-pathway
integration, and advantage-conditioned offline policy improvement.
During pre-training, the policy learns broad contact priors from NeoData, our large-scale visuo-tactile robot dataset. To our knowledge, \modelfull{} is the first VTLA model pretrained on tactile data at scale.
During post-training, we augment the policy with a
predictive tactile pathway,
distilling the contact patterns learned at scale into the fine motion
adjustments in downstream tactile-centric manipulation. 
For offline policy
improvement, we introduce \alter, an advantage-conditioned offline Reinforcement Learning (RL) method that
converts relative progress and trajectory-event comparisons into binary
advantage labels for policy training on a fixed deployment corpus. This
procedure further improves task-specific learning on contact-rich skills such
as deformable object manipulation.
Across contact-rich benchmarks, \modelfull{} outperforms strong baselines
by wide margins: it wins all nine real-robot NeoReal tasks and reaches
$63.8\%$ mean success on the twenty-task simulation suite against $44.0\%$
for the strongest baseline. \model{} policies trained with \alter reach $75$--$95\%$
success on three long-horizon real-robot tasks.
Results
lay a foundation for versatile tactile-driven manipulation policies.

%% file: sections/intro.tex
\section{Introduction}
\label{sec:intro}

Vision--language--action (VLA) models have made manipulation policies
general. Fine-tuned from pretrained vision--language backbones, they follow
instructions and transfer across tasks, scenes, and embodiments
\citep{openvla2024,pi0_2024,pi05_2025}. Touch, however, has remained
largely absent from this progress, leaving current policies with a
persistent weakness in contact-rich manipulation: the tactile extensions
attempted so far train on task-scale collections, at most tens of hours
gathered for a handful of skills.
This report presents, to our knowledge, the first VLA policy pretrained on
tactile data at scale. We scope the report to vision-based
tactile sensing \citep{gelsight2017,digit2020}, instrumenting each gripper finger with our self-developed
sensor so that contact is read out as an image. This signal is nothing like
a camera view: tactile frames are noisy, nearly empty away from contact,
and informative almost only in the brief windows when contact forms or is
about to change.

\begin{figure}[!t]
  \centering
  \includegraphics[width=\linewidth]{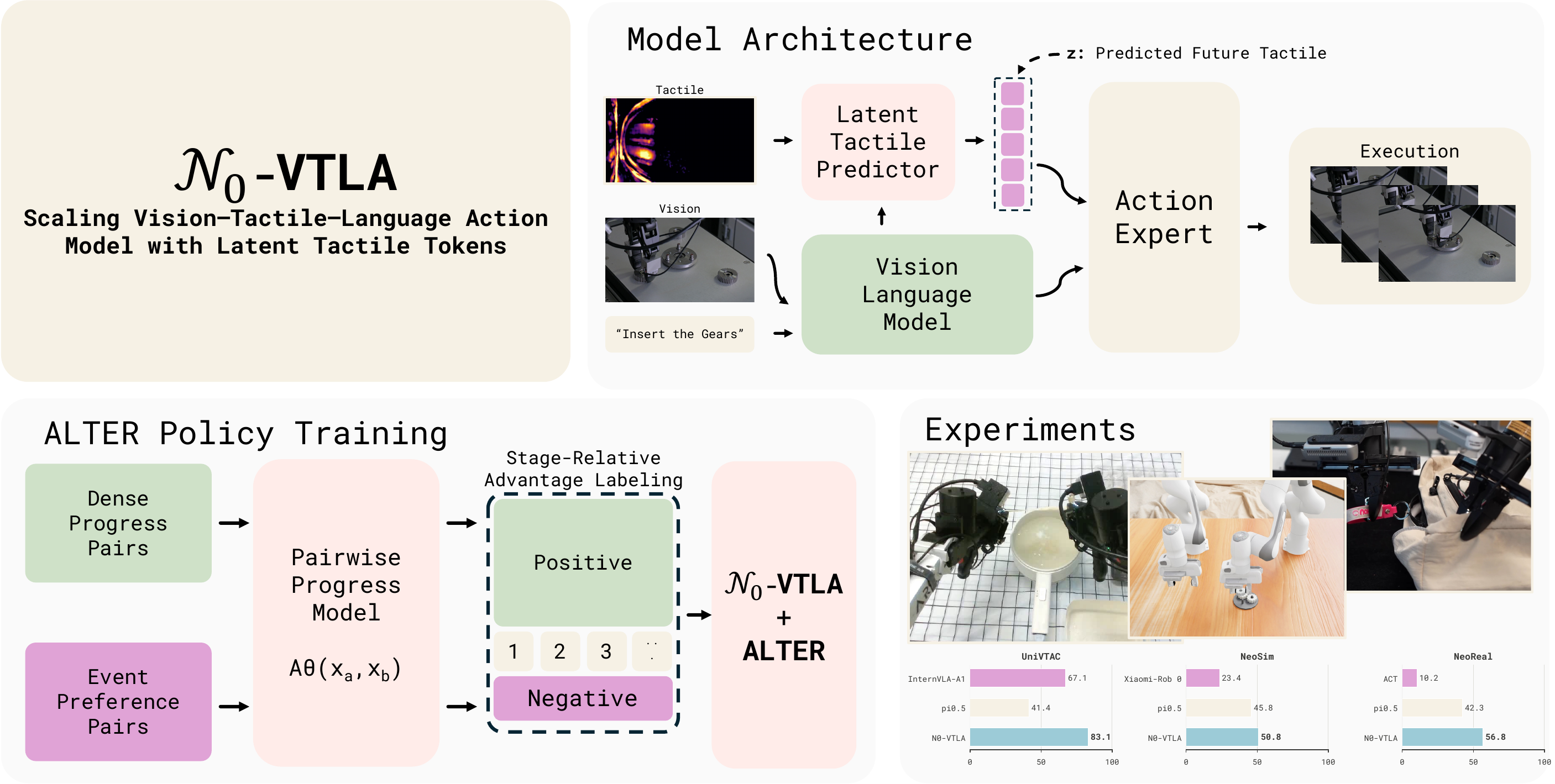}
  \caption{\textbf{\model{} at a glance.} \model{} encodes vision, the
    instruction, and tactile difference images, predicts latent
    tactile tokens $z$, and conditions a flow-matching action expert on
    them. Beyond demonstrations, \alter converts
    deployment experience into stage-relative advantage labels for
    offline policy learning. The experiment panels plot the headline means against
    the base VLA policy and the strongest specialist baseline on UniVTAC
    \citep{univtac2026}, NeoSim, and NeoReal. \model{} leads every panel.}
  \label{fig:overview}
\end{figure}

Existing systems integrate touch along one of two paths. The first
concatenates tactile tokens into the vision--language context and treats
the tactile stream as one more camera \citep{tactilevla2025,omnivtla2025};
yet a signal that is sparse and mostly silent buys little in a prefix built
for information-dense views. The second injects the current tactile reading
into the action pathway to guide denoising \citep{rdp2025,touchguide2026}; this placement mischaracterizes the role of touch, since a tactile frame
records contact that actions already taken have produced and, by itself,
says little about the contact the next actions must anticipate. Conditioned
on it alone, the policy stays one step behind its own contact events.

\model{}, which we read as NeoVTLA, takes a third path: it keeps tactile
out of the vision--language prefix and conditions the action expert on a
prediction of touch rather than the current reading. A small predictor
reads the vision--language context together with the current tactile tokens
and emits \emph{latent tactile tokens} $z$ that estimate the net tactile
change over the coming action chunk, so the policy acts on the contact its
own actions are about to cause. The tactile frames are contact-difference
images encoded by a frozen pretrained visual encoder through a
lightweight trainable projection, and a three-stage recipe brings this
newly initialized pathway online, as Figure~\ref{fig:overview} shows. Beyond
supervised task adaptation, we formulate learning from stored deployment experience as advantage-conditioned offline RL. \alter trains a pairwise progress model
from clean demonstrations, tactile-detected object-drop events, and logged human
corrections, then assigns stage-relative binary advantage conditions for
policy learning.

The full system rests on a vision--language--action backbone built on
PaliGemma \citep{paligemma2024}, a
canonical cross-embodiment action space, and NeoData, a multi-platform
visuo-tactile corpus spanning single and dual-arm configurations,
documented in a companion data report \citep{neoteaidata2026}. Before
evaluating the full system, we verify the latent pathway itself: after
\stageone{}, the latent tokens retrieve their matching future-tactile
targets at $92.3\%$ top-1 accuracy, where chance sits at $3.2\%$. Whether
this grounded representation translates into better task performance is the
more demanding test. We evaluate \model{} against external baselines on the
NeoReal real-robot benchmark and the simulated contact-rich suite under
identical protocols. On the twenty-task simulation suite \model{} leads the
strongest baseline by a wide margin, and it wins all nine real-robot
NeoReal tasks.

In summary, this report makes three contributions:
\begin{itemize}
  \item \textbf{Large-scale tactile pretraining (\S\ref{sec:method:base}).}
        \model{} is pretrained on NeoData, the
        large-scale visuo-tactile robot data across multiple robot platforms, made
        trainable as one model by a canonical cross-embodiment action space
        and a quality-verified data pipeline \citep{neoteaidata2026}.
  \item \textbf{Latent tactile tokens (\S\ref{sec:method:latent}).} Touch is
        treated as a prediction target rather than as observation context, a
        predictor estimating the tactile change over the coming action chunk
        and conditioning the action expert directly, brought online stably by
        a three-stage recipe.
  \item \textbf{Offline policy improvement with \alter
        (\S\ref{sec:training:deployment}).} A pairwise progress model,
        supervised by tactile-grounded stage annotations, tactile-detected
        object-drop events, and logged human corrections, produces stage-relative
        advantage labels for offline policy learning. The method applies to both
        the base VLA policy and \model{}, with \model{}+\alter achieving the
        highest success on all three tasks.
\end{itemize}

%% file: sections/model.tex
\section{Model}
\label{sec:method}

\model{} is a policy for contact-rich manipulation. It reads camera
views, a language instruction, robot state, and touch, and it
generates a chunk of future actions. The model consists of a pretrained
vision--language--action backbone built on PaliGemma \citep{paligemma2024} and one
added component, a latent tactile pathway between perception and action.
The backbone carries the views, instruction, and state in its
vision--language prefix and generates the action chunk with a
flow-matching action expert. In the added pathway, a frozen-backbone
tactile encoder turns each finger's contact-difference image into tokens,
and a small predictor distills those tokens, in the context of the scene
and instruction, into \emph{latent tactile tokens} $z$ that estimate the
net contact change expected over the coming action chunk. The action
expert is conditioned on $z$ directly, and tactile never enters the
vision--language prefix. Touch therefore enters the policy as a
prediction target rather than as one more observation.
Figures~\ref{fig:predictor}--\ref{fig:e2e} lay out this design
as a three-step recipe, and the section follows them.
Section~\ref{sec:method:base} fixes
the base policy, and Section~\ref{sec:method:latent} walks through the
tactile pathway step by step.

\subsection{Base Architecture}
\label{sec:method:base}

The base policy pairs a PaliGemma
vision--language backbone \citep{paligemma2024} with a
flow-matching action expert \citep{flowmatching2022}. Camera views, the
instruction, and the robot state form the model \emph{prefix}, state
entering that prefix in discretized form rather than as a separate
continuous input.
Conditioned on the prefix, the expert denoises an action chunk over a horizon of
$H=50$ steps in the canonical $32$-dimensional container of
Section~\ref{sec:data:canonical}, whose width and slot layout we inherit
unchanged from the pretrained action head so that its
weights load directly. The flow-matching objective is
unmasked over all $32$ dimensions. Each platform populates the dimensions
its embodiment uses, the rest carry zero targets the model learns to
reproduce, and single- and dual-arm data therefore coexist in one model
under a single fixed-width objective. All other aspects, including architecture,
tokenization, and training procedure, are inherited unchanged from
the pretrained backbone.

\subsection{Latent Tactile Tokens}
\label{sec:method:latent}

\begin{figure}[!tp]
  \centering
  \includegraphics[width=.92\linewidth]{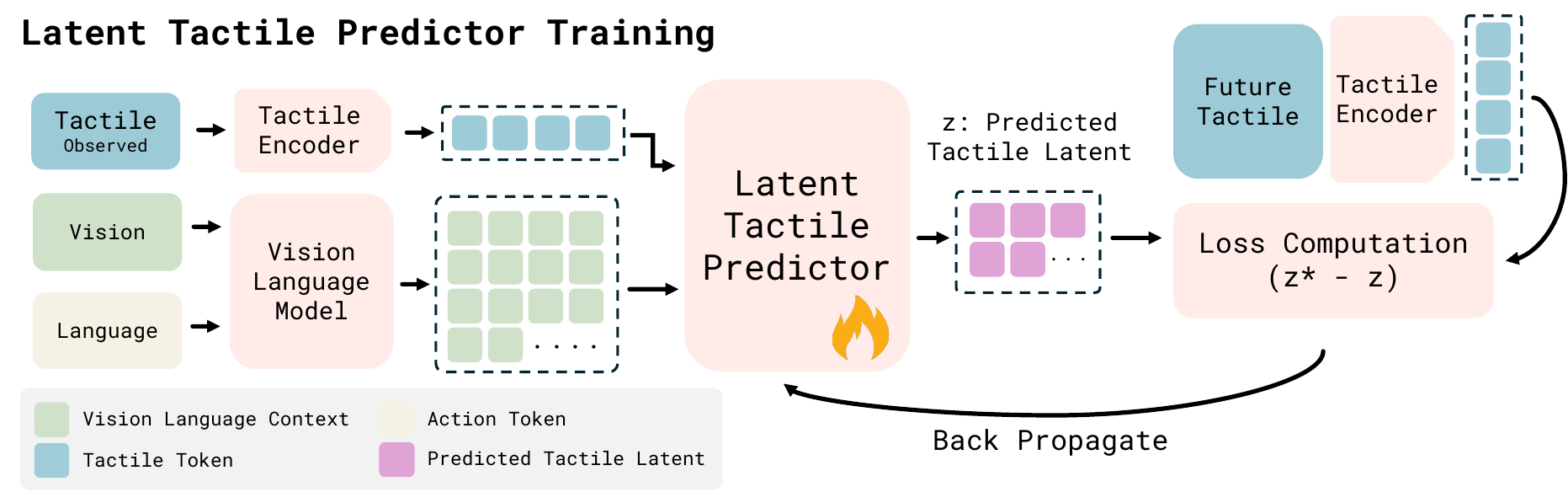}
  \caption{\textbf{Step 1: latent tactile predictor training.} The
    current tactile difference is encoded into tokens $g$. The predictor
    reads $g$ together with the contextualized vision--language prefix and
    emits the latent tactile tokens $z$. The future-tactile target $z^{*}$
    comes from the same tactile encoder applied to the coming tactile
    change, and the loss on $z^{*}$ and $z$ backpropagates into the
    predictor alone.}
  \label{fig:predictor}
\end{figure}

\begin{figure}[!tp]
  \centering
  \includegraphics[width=\linewidth]{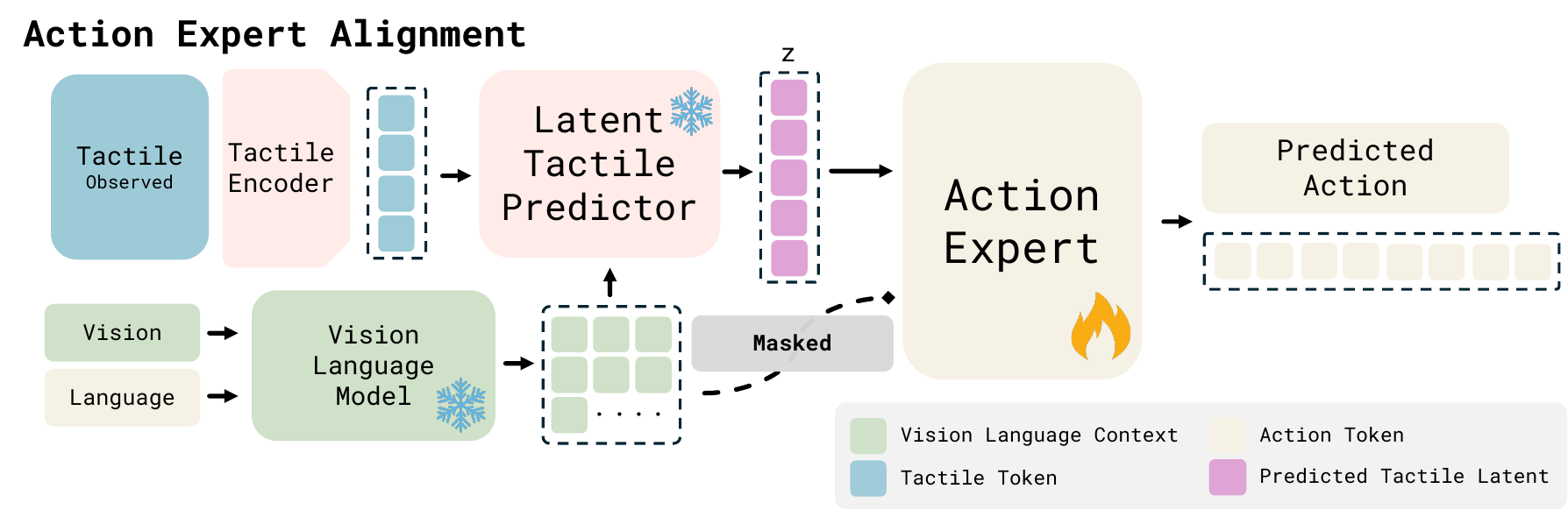}
  \caption{\textbf{Step 2: aligning the action expert with the latent
    tokens.} With the predictor and the vision--language backbone frozen,
    the vision--language context is masked before the action expert, so
    action prediction must draw on the latent tactile tokens $z$ while the
    expert learns the interface.}
  \label{fig:aea}
\end{figure}

Figure~\ref{fig:predictor} shows Step~1, in which the tactile predictor is
trained against a future-tactile target. Figures~\ref{fig:aea}
and~\ref{fig:e2e} show
Steps~2 and~3, in which the action expert first learns to consume the resulting
latents while the vision--language pathway is masked, and the full policy
then trains end to end. The model that leaves Step~3 is the deployed
controller. The paragraphs below introduce each component in the same
order.

\paragraph{The tactile encoder.}
\label{sec:method:tactile}
Every panel of both figures begins the same way. The policy never sees a raw
tactile frame. For view $k$ we subtract the episode-start baseline frame
$\mathrm{tac}_0^{k}$, the zero-contact reference established in
Section~\ref{sec:data:tactile}, from the current frame
$\mathrm{tac}_{\tau}^{k}$ in pixel space, and encode the difference with a
frozen self-supervised visual encoder,
followed by a trainable linear projection to the shared
token width $d$ of the vision--language backbone:
\begin{equation}
  g_k = f_{\mathrm{enc}}\!\big(\mathrm{tac}_{\tau}^{k} - \mathrm{tac}_0^{k}\big)
        \in \mathbb{R}^{10\times d},
  \label{eq:g}
\end{equation}
where $f_{\mathrm{enc}}$ denotes the frozen encoder composed with the
trainable projection. Each tactile image yields $10$ tokens, one class token and nine
spatial tokens from a $3\times3$ adaptive average pool over the
encoder's $16\times16$ patch grid, and the tokens of the $n$ active views are
concatenated into $g = [\,g_1;\dots;g_n\,]\in\mathbb{R}^{10n\times d}$.
Differencing against a per-episode baseline, rather than encoding the
absolute gel image, removes the static gel appearance and much of the
mount-specific imprint, making the representation robust to, though not
strictly invariant under, differences in sensor placement. Freezing the
encoder is deliberate. It preserves the self-supervised representation
intact, it lets a previously unseen sensor be onboarded by training only
the lightweight projection, and it removes the encoder's activations and
optimizer state from the training memory budget.

\paragraph{Step 1: the predictor and its future-tactile target.}
The predictor is a lightweight module that reads the current tactile
difference tokens $g$ in the context of the scene and instruction, carried
by the contextualized vision--language prefix, and distills them into a
compact set of learned latent queries that become the latent tactile tokens
$z$. When an episode carries no tactile at all, a learned null token stands
in for $g$, so $z$ is always produced and the policy falls back to
vision--language control rather than failing. What makes $z$
predictive rather than merely descriptive is the target
Figure~\ref{fig:predictor} attaches to it,
\begin{equation}
  z^{*} = \frac{1}{n}\sum_{k=1}^{n}
          f_{\mathrm{enc}}\!\big(\mathrm{tac}^{k}_{\tau+H} - \mathrm{tac}^{k}_{\tau}\big)
          \in \mathbb{R}^{10\times d},
  \qquad H = 50,
  \label{eq:zstar}
\end{equation}
obtained by applying the same tactile encoder to each view's tactile
change over the next $H$ steps and averaging the resulting tokens across
the $n$ active views. The predictor output $z\in\mathbb{R}^{10\times d}$
is trained to match $z^{*}$. The supervision combines a symmetric InfoNCE\citep{infonce2018}
contrastive loss that pulls the predicted latent toward its matching
future-tactile target and an auxiliary $L_1$ reconstruction of a coarse
future-tactile-difference field. This is the supervision with which the three-stage recipe
grounds the predictor. A
\emph{free-latent} simplified variant omits the Step~1 supervision
entirely, shaping $z$ through the action gradient alone. It requires no
future frames during training.

\paragraph{Step 2: conditioning the action expert.}
The latent tokens $z$ are projected to the action-expert width and
prepended to the action suffix, ahead of the noisy action tokens.
The current-contact tokens $g$ never enter the action expert. They reach
action generation only through the predictor, which distills them into
$z$. The latent tokens form their own conditioning block. The action
tokens attend to $z$ and, as in the base policy, to the vision--language
prefix, the prefix never attends back to the latent tokens, and the $z$
positions are sliced off before the action output head so that they never
emit actions. Because the pretrained expert has never consumed such a
token, Step~2 (Figure~\ref{fig:aea}) trains this interface in
isolation. The
vision--language pathway is masked so that action prediction must draw on
$z$, aligning the latents with the expert before anything else moves.

\paragraph{Step 3: training the full policy end to end.}
Step~3 (Figure~\ref{fig:e2e}) removes that mask and opens the whole policy
to joint training. The direct prefix-to-expert path is restored, so the
expert again sees scene and instruction alongside $z$. Every component
except the frozen tactile encoder backbone then adapts under the action
objective. What is frozen at each step, and why the order matters, is the
subject of the training chapter.

\begin{keybox}
\textbf{Why predict, not react?} The design choice is what conditions the
action expert. Handing the policy its current tactile reading, whether
concatenated into the prefix or injected alongside the actions, hands it
a record of contact already made, and spends capacity built for
information-dense views on a signal that is sparse and mostly silent. The
latent predictor instead asks the model to form an explicit internal
estimate of the contact state, and, under supervision, of the net contact
change expected over the chunk horizon. It
then conditions the action head on that estimate. The behaviors where
tactile matters most are anticipatory, such as the millimeter-scale pre-load
before a grasp closes and the catch of incipient slip before the object
moves. These live in the pre-contact blind spot, where a purely reactive
signal arrives too late to shape the action that caused it. We therefore
hypothesize that conditioning actions on a predictive latent, rather than
on raw current tokens, is the better inductive bias for contact-rich
control. This choice to predict in a learned latent space rather than
reconstruct raw sensory signals follows the joint-embedding predictive
principle \citep{ijepa2023} explored for self-supervised representation learning in LeJEPA
\citep{lejepa2025} and for latent world modeling from pixels in
LeWorldModel \citep{leworldmodel2026}.
\end{keybox}

Two observations ground this design choice. What decides a $50$-step
chunk is the contact the chunk itself is about to create, and no encoding
of the present frame contains it. Ranking future-tactile targets by the
current tactile encoding alone retrieves $57\%$ top-1 where the predictor
reaches $92.3\%$, with the margin widening as the candidate pool grows, as
Section~\ref{sec:experiments:repr} details. And the prediction
objective pins the latent to touch before the policy ever optimizes
through it, where features shaped by the action gradient alone would be
free to drift into an appearance cue rather than contact state.

\FloatBarrier

%% file: sections/data.tex
\begin{figure}[!tp]
  \centering
  \includegraphics[width=\linewidth, trim=0 0 0 22, clip]{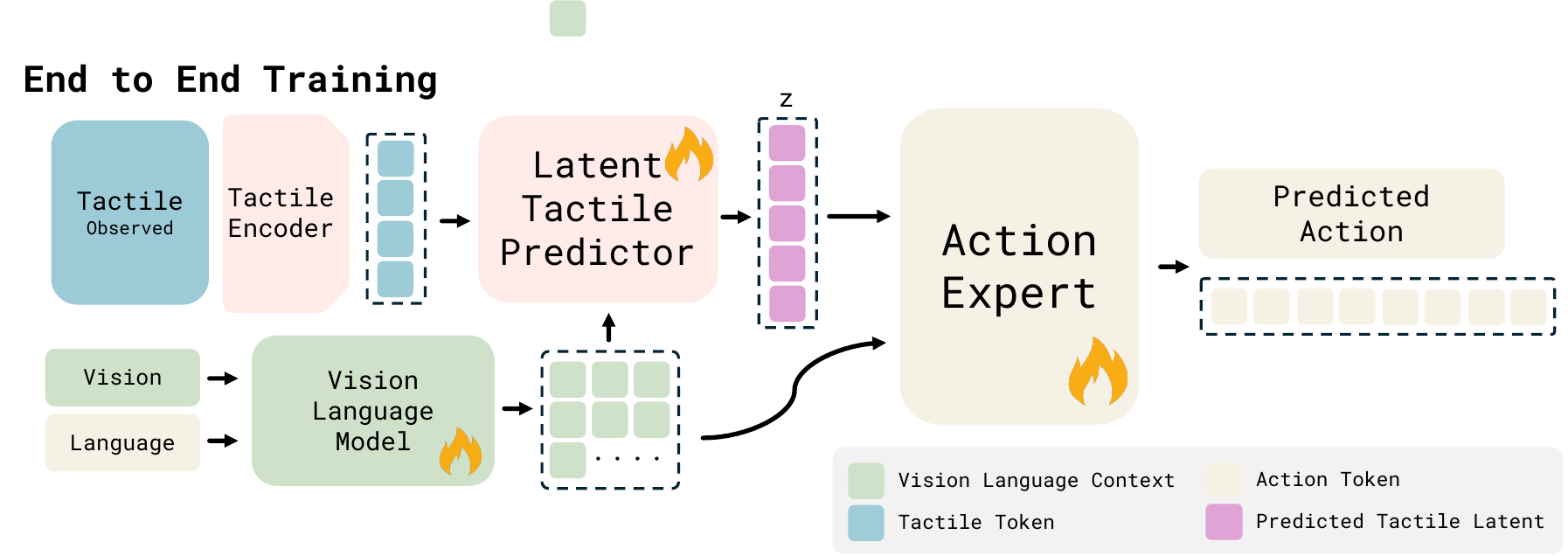}
  \caption{\textbf{Step 3: end-to-end training.} Everything except the
    tactile encoder backbone unfreezes and the full policy trains end to
    end, emitting the predicted action chunk. The free-latent simplified
    variant omits the Step~1 supervision, shaping $z$ through the action
    gradient alone.}
  \label{fig:e2e}
\end{figure}

\section{Data}
\label{sec:data}

\model{} is pretrained on NeoData~\citep{neoteaidata2026}, our large-scale curated
multi-platform visuo-tactile corpus,
spanning single- and dual-arm robot manipulators as well as a UMI-style
handheld collection gripper \citep{umi2024}. Every gripper finger that participates in a manipulation task carries
our self-developed visuo-tactile sensor,
so contact is read out as a stream of tactile images rather than as a
low-dimensional force signal. Collection protocols, corpus
composition, and sensor specifications are documented in the companion
data report \citep{neoteaidata2026}. This section states only the
conventions the rest of the report depends
on.

\paragraph{Canonical action and state schema.}
\label{sec:data:canonical}
All embodiments are unified into one fixed $32$-dimensional state and
action container, inherited from \pizerofive{}
\citep{pi05_2025}. The container is laid out for two arms, with the first
$20$ dimensions split into one $10$-dimensional slot per arm and the
remaining $12$ left unused and always zero. Within a slot, the $10$
dimensions comprise a $3$-dimensional end-effector position, a
$6$-dimensional rot6d rotation \citep{zhou2019rot6d}, and a
$1$-dimensional gripper channel. Dual-arm episodes populate both slots.
Single-arm episodes populate the first slot only and leave the second
zero-filled as well. Single- and dual-arm data therefore coexist
in one fixed-width container, and what the policy does with the
zero-filled dimensions is fixed by the objective of
Section~\ref{sec:method:base}. Actions are stored as absolute
end-effector poses. At training time each chunk is rewritten relative to
its own first frame, so the model predicts motion relative to the pose at
which the chunk begins. At deployment the predicted chunk is mapped back
to absolute poses through the inverse of that transform, and inverse
kinematics resolves those poses into the joint commands sent to the
robot.\label{sec:data:norm} Normalization statistics are computed on the
chunk-relative representation, separately for each pairing of robot and
action schema.

\paragraph{Tactile collection convention.}
\label{sec:data:tactile}
Each participating gripper finger contributes one tactile stream,
captured on the same clock as the RGB and proprioceptive channels. The
per-platform stream counts and the common frame rate are listed in the
data card of Appendix~\ref{app:datacard}. Each episode begins with a
short zero-contact baseline, the gripper open and static for at least
$0.5$\,s. That baseline frame is the episode's zero-contact reference, and
every tactile frame recorded afterwards is interpreted relative to it.

\paragraph{Data quality verification.}
\label{sec:data:gates}
Every converted repository is verified for data quality before it enters
training. Verification checks that a repository is complete and
internally valid, that its stored conventions match the schema above, and
that its statistics and media are consistent with what the training
pipeline assumes. Repositories that fail are repaired or excluded rather
than trained on. The individual invariants, and the symptom each produces
when it is violated, are catalogued in Appendix~\ref{app:pitfalls}.

\paragraph{Simulated data.}
\label{sec:data:sim}
Simulated data enters through the same door. Episodes from the UniVTAC
visuo-tactile simulator \citep{univtac2026}, which supplies the NeoSim
suite evaluated in Section~\ref{sec:experiments:neosim}, are converted
into the canonical schema above and verified alongside real data, so that
one policy interface applies to both.

%% file: sections/training.tex
\section{Training}
\label{sec:training}

\model{} reaches deployment through three core phases. A three-stage recipe then
brings the latent tactile pathway
online. Supervised post-training specializes the resulting generalist to
individual tasks. After this core recipe, an optional procedure
post-trains the task policy on its own deployment data. Throughout, the
trainable surface grows only after each new interface has been grounded,
so the tactile pathway comes online without destabilizing the pretrained
policy.

\subsection{Base Pre-training}
\label{sec:training:pretrain}

\begin{figure}[!tp]
  \centering
  \includegraphics[width=.85\linewidth]{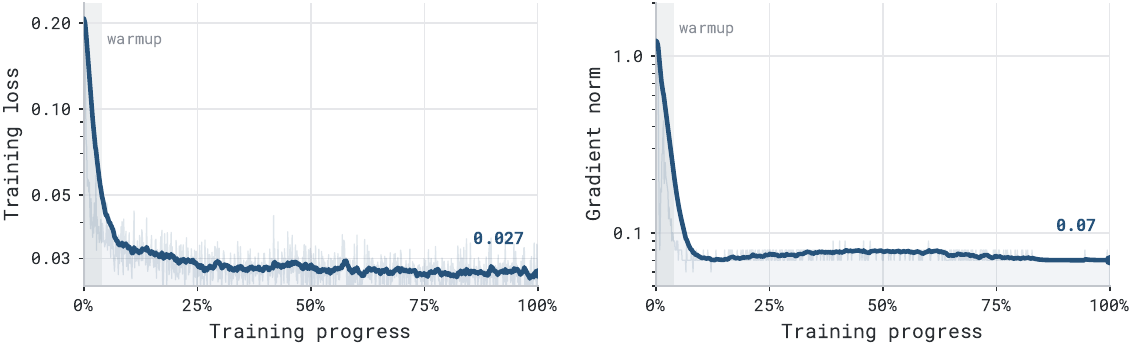}
  \caption{\textbf{Multi-platform visuo-tactile pretraining.} Training loss and
    gradient norm over multi-platform visuo-tactile pretraining. The loss
    descends smoothly and gradient norms stay flat throughout, consistent with
    the pretrained initialization transferring cleanly to the visuo-tactile
    action space.}
  \label{fig:pretrainloss}
\end{figure}

Base pre-training is conducted at cluster scale on the NeoData corpus.
Stability at this scale is achieved by design, through choices validated
in controlled comparisons, and is visible in the smooth loss and flat
gradient norms of Figure~\ref{fig:pretrainloss}.

\subsection{Three-Stage Latent-Tactile Training}
\label{sec:training:stages}

The tactile pathway, newly initialized, attaches to the pretrained
multi-platform checkpoint, and the three steps of
Figures~\ref{fig:predictor}--\ref{fig:e2e} bring it online,
each stage proceeding from the checkpoint the previous one
produces. The free-latent configuration
corresponds to collapsing this recipe into a single joint stage with no
auxiliary supervision.

\paragraph{\stageone: grounding the predictor.}
With the entire base policy frozen, we train only the predictor, the
tactile projection, and a lightweight reconstruction head. The predictor
output $z$ is pulled toward the future-tactile target $z^{*}$ of
Eq.~\ref{eq:zstar} by a symmetric InfoNCE\citep{infonce2018} objective. Write $h(\cdot)$ for the mean pooling over the ten latent tokens followed by $\ell_2$ normalization, and, for a batch of $B$ samples,
\begin{equation}
  s_{ij} = \big\langle h(z_i),\,h(z^{*}_j)\big\rangle
  \label{eq:sim}
\end{equation}
for the cosine similarity between the $i$-th
prediction and the $j$-th target. The contrastive term
\begin{equation}
  \mathcal{L}_{\mathrm{NCE}}
    = -\frac{1}{2B}\sum_{i=1}^{B}
      \left[
        \log\frac{e^{s_{ii}}}{\sum_{j=1}^{B} e^{s_{ij}}}
        + \log\frac{e^{s_{ii}}}{\sum_{j=1}^{B} e^{s_{ji}}}
      \right]
  \label{eq:infonce}
\end{equation}
matches each predicted latent to its own future target, taking the other
targets in the batch as negatives, and is symmetrized over both retrieval
directions. In parallel a reconstruction head $r_\psi$ decodes $z$ back to
a coarse future-tactile-difference field and is trained with an $\ell_1$
term $\mathcal{L}_{\mathrm{rec}} = \lVert r_\psi(z) -
\bar{D}_{\tau\rightarrow\tau+H}\rVert_1$ against the same horizon, where
$\bar{D}_{\tau\rightarrow\tau+H}$ is the downsampled contact-change field
over the coming chunk. The stage minimizes
\begin{equation}
  \mathcal{L}_{1} = \mathcal{L}_{\mathrm{NCE}}
    + \lambda_{\mathrm{rec}}\,\mathcal{L}_{\mathrm{rec}},
  \label{eq:stage1}
\end{equation}
where $\lambda_{\mathrm{rec}}>0$ balances the two terms.
The contrastive term supplies the discriminative pressure that makes $z$
retrieve the right future contact; the reconstruction term anchors it to
the spatial layout of that contact, discouraging a shortcut latent that
separates batches without encoding where contact forms. Because gradients
touch only the shallow predictor stack, the stage is cheap and converges
quickly. At convergence the latent tokens are strongly grounded in touch:
the latent $z$ retrieves its matching future-tactile target with $92.3\%$
top-1 accuracy against a $3.2\%$ random baseline, analyzed in
Section~\ref{sec:experiments}.

\paragraph{\stagetwo: aligning latents with the action expert.}
The pretrained action expert has never consumed a latent tactile token, so
we next teach it the interface. We hold the tactile perception stack
frozen at its \stageone{} checkpoint and train only the latent-to-expert
projection and the action expert, under the base action objective.
Concretely, in the expert's attention the keys and values from the
vision--language prefix are masked out for the action queries, leaving the
latent tokens $z$ and the noised action tokens as the only conditioning
the expert can attend to. Masking the prefix removes the shortcut of
predicting actions from scene and instruction alone, so the only route to
lowering the action loss runs through $z$. The expert learns to read touch
through $z$ before any joint training loosens the rest of the policy, in
the spirit of the staged alignment strategies explored for
language--action models \citep{qwenvla2026}.

\paragraph{\stagethree: end-to-end joint training.}
With the predictor grounded and its interface aligned, we unfreeze
everything except the always-frozen tactile encoder backbone and train the
full policy jointly under the standard pre-training recipe, on the action
objective alone. The vision--language mask of \stagetwo{} is removed, so
the direct prefix-to-expert path is restored and the expert again sees
scene and instruction alongside $z$. Gradients from the action objective
now flow together through the predictor, the two projections, the action
expert, and the vision--language backbone, letting the perception stack
adapt to what the expert actually needs. The contrastive and
reconstruction targets of \stageone{} are no longer applied. The predictor
keeps its grounding through the action gradient alone, which the
perturbation probe of Section~\ref{sec:experiments:repr} confirms it
retains rather than reroutes.

\begin{keybox}
\textbf{Why three stages?} The staging turns one hard joint optimization
into a curriculum, each stage handing the next a better starting point.
\stageone{} fixes what $z$ \emph{means}: trained only against the
future-tactile target, the bottleneck is forced to encode contact rather
than a second copy of the prefix it already sees. \stagetwo{} fixes how
the expert \emph{uses} $z$: with the prefix masked, the only way to lower
the action loss is to read touch through $z$, so the expert commits to the
new interface while the rest of the policy stays put. \stagethree{} then
trains everything jointly, but from an initialization where $z$ is already
grounded and already consumed, so joint optimization refines a working
pathway instead of having to discover grounding and interface at once. The
free-latent variant collapses all three into a single joint stage: with no
\stageone{} target, the action gradient alone shapes $z$, and nothing stops
the bottleneck from degenerating into a copy of the prefix or being
bypassed altogether. After joint training the latent stays tactile-leaning
(Section~\ref{sec:experiments:repr}), consistent with \stagethree{}
refining the grounded pathway rather than rerouting around it. A randomly
initialized tactile pathway can therefore be grafted onto a pretrained VLA
and brought online without destabilizing it.
\end{keybox}

\subsection{Supervised Task Adaptation}
\label{sec:training:sft}

We adapt a pretrained \model{} checkpoint to each downstream task by
supervised fine-tuning(SFT) on a few hundred demonstrations, warm-starting
from that checkpoint and reusing the pre-training recipe at reduced
scale. Normalization statistics are always recomputed on
the task's own data and never reused from pre-training, per
Section~\ref{sec:data:norm}. The same recipe covers both real-robot tasks
and the simulated task suite, with one policy per task.

\subsection{Offline RL from Deployment Data with \alter}
\label{sec:policyimp}
\label{sec:training:deployment}

\begin{figure}[!tp]
  \centering
  \includegraphics[width=\linewidth]{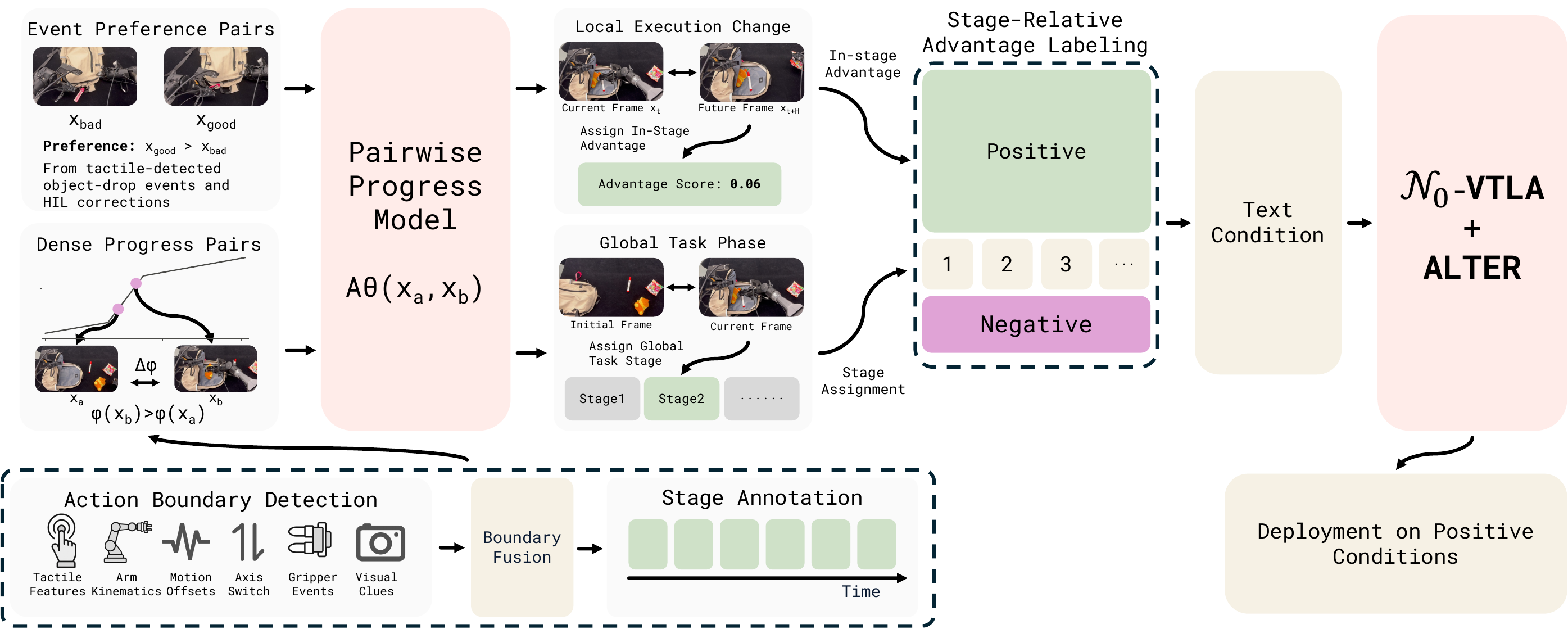}
  \caption{\textbf{\alter for offline policy improvement.}
  Tactile contact changes, complemented by kinematic and visual cues, ground
  stage annotations of clean demonstrations and yield dense progress pairs.
  Tactile-detected object-drop events and logged HIL corrections yield sparse preference pairs. Both supervise a pairwise progress model, which is then frozen to
  estimate global task phase and local execution change for each stored
  trajectory. Within each predicted stage, local-change estimates produce
  binary advantage labels that are appended to the task prompt during policy
  training. At deployment, the task prompt uses the positive label.}
  \label{fig:pimethod}
\end{figure}

We call our method \alter, short for Advantage Labeling from Trajectory Events and Relative Progress. Given a fixed deployment corpus, \alter performs advantage-conditioned offline RL \citep{pistar06_2025} without additional environment interaction,
as summarized in Figure~\ref{fig:pimethod}. Clean demonstrations
provide dense progress supervision from signal-grounded stage intervals, whose boundaries are localized using tactile contact changes, end-effector
kinematics, gripper state, and visual event cues. Imperfect deployment
trajectories instead provide sparse before-and-after preferences from tactile-detected object-drop events and logged human-in-the-loop corrections.
These dense and sparse signals jointly train a task-specific pairwise progress model. We then freeze
the model and apply it to every trajectory retained for offline policy learning.
Comparing each observation with the episode start estimates global task phase,
while comparing it with the observation one action-chunk later estimates local
execution change.
Within each predicted stage, we rank samples by estimated local change, assign
positive labels to higher-ranked samples and negative labels to lower-ranked
ones, and append the resulting label to the task prompt used for policy
training.

\paragraph{Deployment corpus and offline annotations.}
\label{sec:policyimp:data}
\label{sec:policyimp:labels}

After deploying task-adapted policies, we retain clean teleoperation
demonstrations, autonomous rollouts, and human-in-the-loop (HIL) rollouts.
The HIL corpus also includes staged-recovery episodes that start from selected
error states and record human-teleoperated recovery trajectories.
Clean demonstrations receive dense stage-progress annotations from a
signal-grounded pipeline. From representative demonstrations,
Gemini-3.5-Flash\citep{gemini35flash} generates a shared task template comprising the L3
objective, ordered L2 stages, and their L1 steps. A human reviews this
task-level template once, after which its vocabulary and ordering remain fixed
across demonstrations of that task. For each demonstration, tactile contact
changes identify candidate transition times, supplemented by end-effector
motion, gripper state, and visual GEBD cues \citep{shou2021gebd}. After these
candidates are merged and deduplicated, the VLM maps each resulting interval to
an entry in the task template. It therefore assigns stage semantics to
pre-segmented intervals rather than predicting timestamps from the full video.
We do not assign monotone labels to complete
autonomous or HIL trajectories because they may regress or retry.
Instead, tactile contact loss localizes object-drop events, while HIL logs mark
correction intervals. These timestamps define local event comparisons.
Appendix~\ref{app:valueablation} reports the task-wise composition
of this corpus and provides annotation examples.

\paragraph{Duration-calibrated stage progress.}
For each stage $k$, we first compute its mean duration $\bar d_k$ across clean
demonstrations. Its share of the task progress is
\begin{equation}
  w_k=\frac{\bar d_k}{\sum_{j=1}^{K}\bar d_j},
  \qquad
  \phi_t=\sum_{j<k}w_j+w_k u_t.
  \label{eq:pi-duration-progress}
\end{equation}
Here $K$ is the number of stages, and $u_t\in[0,1]$ is the fraction of the
current stage completed at frame $t$. Thus $\phi_t$ combines the cumulative
weights of completed stages with the duration-scaled fraction of the current
stage.
Weighting stages by their mean durations avoids forcing a brief transition
and a long manipulation stage to occupy equal portions of the $[0,1]$ range.

\paragraph{Event-aware pairwise progress learning.}
\label{sec:policyimp:pairs}

We implement $A_\theta(x_a,x_b)\in[-1,1]$ with a \pizerofive{}-based
paired-observation progress architecture \citep{chi0_2026}. Conditioned on
the task prompt, the model jointly encodes observations from two time points,
each represented by synchronized RGB images from multiple cameras. A
three-layer MLP head then predicts the relative task progress between them.
Tactile, kinematic, and event signals construct the offline targets but are not
inputs to the progress model.
For demonstration pairs, we regress $A_\theta(x_a,x_b)$ toward the target
$\phi(x_a)-\phi(x_b)$. We label an observation immediately before an object
drop as higher-progress than one immediately after it. For a HIL correction,
we label an observation near the end of the intervention as higher-progress
than one near its start. Let $\mathcal D_{\mathrm{event}}$ contain triples
$(x_a,x_b,y)$, where $y=+1$ indicates that $x_a$ is assigned higher progress
than $x_b$, and $y=-1$ indicates the reverse. We optimize
\begin{equation}
  \begin{aligned}
  \mathcal L_{\mathrm{prog}}
  ={}&\mathbb E_{\mathcal D_{\mathrm{stage}}}
  \left[A_\theta(x_a,x_b)-(\phi(x_a)-\phi(x_b))\right]^2\\
  &+\lambda_{\mathrm{event}}
  \mathbb E_{\mathcal D_{\mathrm{event}}}
  \left[\max(0,m-yA_\theta(x_a,x_b))\right]^2.
  \end{aligned}
  \label{eq:pi-event-objective}
\end{equation}

Here $\mathcal D_{\mathrm{stage}}$ contains demonstration pairs with dense
progress-difference targets, $m>0$ specifies the minimum signed score
$yA_\theta(x_a,x_b)$ required of an event pair, and
$\lambda_{\mathrm{event}}$ controls the overall contribution of event
comparisons relative to dense demonstration supervision.
We randomly reverse event-pair input order during sampling to prevent a fixed
input slot from becoming a shortcut.

\paragraph{Advantage-conditioned offline policy learning.}
\label{sec:policyimp:acp}
For every stored episode, the frozen pairwise progress model produces global
progress
$\hat\phi_t$ and local change $\hat r_t$. Samples are assigned to the
duration-calibrated stage containing $\hat\phi_t$, then ranked only against
other samples from that stage:
\begin{equation}
  \begin{aligned}
  \hat\phi_t&=\delta_e+A_\theta(x_t,x^e_0), &
  \hat r_t&=A_\theta(x_{\min(t+H,T_e-1)},x_t),\\
  q_k&=\operatorname{Quantile}_{1-\rho}\{\hat r_i:s_i=k\}, &
  c_t&=\mathbf 1[\hat r_t\geq q_{s_t}].
  \end{aligned}
  \label{eq:pi-advantage-condition}
\end{equation}
Here $x^e_0$ is the first observation of episode $e$, $T_e$ is its number of
frames. The offset $\delta_e$ is zero for episodes that begin at the nominal
task start; staged-recovery episodes use the initialization described in
Appendix~\ref{app:valueablation}. The estimated progress $\hat\phi_t$
determines stage index $s_t$, and $q_k$ is the within-stage threshold. We use
the action-chunk horizon $H=50$ and retain the top $\rho=0.3$ within each stage
as positive, as indicated by $c_t=1$. We represent this binary indicator as an
additional text input. Samples with $c_t=1$ receive
\texttt{Advantage: positive}, while the remaining samples receive
\texttt{Advantage: negative}. The tag is appended to the existing task prompt.
For each base policy, \alter training starts from its pretrained
checkpoint.
The policy architecture and training objective remain unchanged, including
the same flow-matching loss used for supervised task adaptation. At deployment
the prompt always uses \texttt{Advantage: positive}. Filtering, recovery offsets,
short-horizon normalization, and sampling hyperparameters are reported in
Appendix~\ref{app:valueablation}.

\FloatBarrier

%% file: sections/experiments.tex
\section{Experiments}
\label{sec:experiments}

We evaluate \model{} on the NeoReal real-robot benchmark and the
twenty-task simulation suite, test further improvement from deployment
data, and close with analyses of the learned representation.
Throughout, the base policy of Section~\ref{sec:method:base} is
initialized from the released \pizerofive{} weights \citep{pi05_2025},
and the frozen tactile encoder of Section~\ref{sec:method:tactile} is
DINOv2 \citep{dinov2_2023}.

\subsection{Evaluation Protocol}
\label{sec:experiments:protocol}

Every comparison in this report is decided by rollout success
rate on the target hardware or simulator. 
Alongside binary success we
report a $100$-point progress score that awards partial credit for
intermediate milestones. Each task is decomposed into a fixed sequence of
subtask checkpoints, a rollout earns credit for the deepest checkpoint it
reaches, and near-miss behavior on contact-rich tasks stays visible.
Representative rubrics appear in Appendix~\ref{app:pertask}, with the
full set in the companion data report \citep{neoteaidata2026}.

On NeoReal, each task carries its own trial budget, and outcomes are
reported as success rate in percent together with the
$100$-point stage-rubric progress score. On NeoSim, each policy is trained
on $100$ demonstrations and evaluated as a success percentage under the
simulator's task-completion criterion with a fixed per-task language
prompt. Comparisons against the no-tactile base policy are isolated
structurally. With its tactile flag disabled, the model reduces to the
base policy, so any measured difference in success rate is attributable
to the tactile pathway alone.

\subsection{Real-World Results: NeoReal}
\label{sec:experiments:neoreal}

We evaluate \model{} on nine tasks from NeoReal, a real-world benchmark of fine-grained contact-rich manipulation tasks defined in the \mbox{$\mathcal{N}_0$}-Foundation data report \citep{neoteaidata2026}.
Tasks run on the
corpus's robot-arm platforms under the shared protocol. 
Figure~\ref{fig:rollouts} shows representative rollouts.
We deploy the post-trained
checkpoints of \model{} and compare against ACT \citep{act2023} and
\pizerofive{} \citep{pi05_2025}, both
reproduced internally under one aligned evaluation protocol.

\begin{figure}[!tp]
  \centering
  \includegraphics[width=\linewidth]{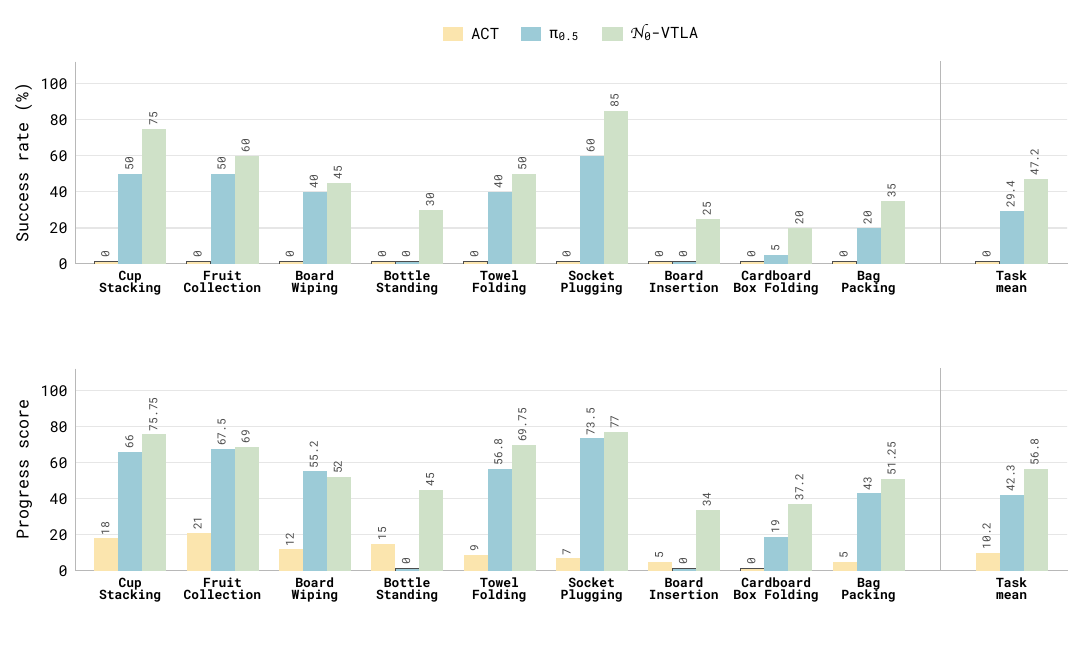}
  \caption{\textbf{NeoReal benchmark: real-world results.} Simulation 
    success rate in the upper panel and the $100$-point progress score in the
    lower panel, on the nine NeoReal contact-rich tasks for ACT,
    \pizerofive{}, and \model{}, with exact values printed above each bar
    and the nine-task means in the rightmost group. Measured zeros appear as
    thin baseline ticks. \model{} beats the strongest baseline on every task
    in success rate and leads the progress-score mean.}
  \label{fig:neoreal}
\end{figure}

Figure~\ref{fig:neoreal} plots per-task success rate and progress score.
\model{} beats the strongest baseline on every task in success rate,
averaging $47.2\%$ against $29.4\%$ for \pizerofive{}. On the progress
score it leads on eight of the nine tasks and in the mean, $56.8$ points
against $42.3$. ACT completes no task and
averages $10.2$ progress points, stalling in the earliest checkpoints. The
margin is clearest on Socket Plugging, the precision outlet
insertion, where \model{}
reaches $85\%$ against $60\%$ for \pizerofive{}, and its successful
rollouts hold up under daylight lighting shifts and recover from failed
insertion attempts. On the long-horizon tasks the progress score separates
the systems more sharply than the success rate alone. On Cardboard Box Folding,
\model{} earns $37.2$ points of stage credit at a $20\%$ success rate,
against $19$ points for \pizerofive{}.

\subsection{Simulation Results: UniVTAC and NeoSim}
\label{sec:experiments:neosim}

The simulation evaluation covers $20$ fine-grained contact-rich tasks on
the UniVTAC framework, the eight original UniVTAC tasks
\citep{univtac2026} and the twelve NeoSim tasks of the companion report
\citep{neoteaidata2026}, four single-arm and eight dual-arm. Both suites
stress the pre-contact and in-contact regimes where an anticipatory
tactile signal should matter most. We report per-task success rate in
percent for
\model{} alongside external baselines, namely
\pizerofive{} \citep{pi05_2025},
StarVLA-$\alpha$ \citep{ye2026starvla}, InternVLA-A1 \citep{cai2026internvla},
Xiaomi-Robotics-0 \citep{cai2026xiaomirobotics}, and
GigaWorld-Policy \citep{ye2026gigaworld}, all evaluated under one
aligned protocol on the same task configurations.

\begin{table}[!tp]
  \centering
  \caption{\textbf{UniVTAC benchmark: per-task success rate (\%).} Closed-loop
    success on the eight UniVTAC tasks for \model{} and external baselines,
    all under one aligned protocol. ACT (Vision Only) drops the tactile
    stream; ACT + UniVTAC and VITaL \citep{vital2024} add it. Best per task
    in \textbf{bold}.}
  \label{tab:univtac}
  \small
  \setlength{\tabcolsep}{3pt}
  \renewcommand{\arraystretch}{1.15}
  \begin{tabularx}{\linewidth}{@{}l*{9}{>{\centering\arraybackslash}X}@{}}
    \toprule
    \multirow{2}{*}{Method}
      & Lift & Pull-out & Lift & Put & Insert & Insert & Insert & Grasp
      & \multirow{2}{*}{Avg}\\
      & Bottle & Key & Can & Bottle & Hole & HDMI & Tube & Classify & \\
    \midrule
    ACT (Vision Only) & 42 & 28 & 20 & 28 & 19 & 15 & 45 & 50 & 30.9\\
    ACT + UniVTAC     & 71 & 46 & 29 & 31 & 24 & 28 & 56 & 99 & 48.0\\
    VITaL             & 72 & 47 & 8  & 32 & 25 & 6  & 34 & \textbf{100} & 40.5\\
    \pizerofive{}     & \textbf{100} & 35 & 6 & 34 & 25 & 8 & 74 & 49 & 41.4\\
    StarVLA-$\alpha$  & 32 & 51 & 65 & \textbf{88} & 52 & 24 & 69 & 68 & 56.1\\
    InternVLA-A1      & 58 & 66 & \textbf{90} & 37 & 93 & 12 & 95 & 86 & 67.1\\
    Xiaomi-Robotics-0 & 21 & 80 & 13 & 12 & \textbf{96} & \textbf{69} & 98 & 45 & 54.3\\
    GigaWorld-Policy  & 38 & 32 & 0 & 21 & 12 & 0 & 9 & 20 & 16.5\\
    \arrayrulecolor{black!30}\midrule\arrayrulecolor{black}
    \rowcolor{metablue!12}
    \modelbf & 99 & \textbf{99} & 88 & 60 & 95 & 25 & \textbf{99} & \textbf{100} & \textbf{83.1}\\
    \bottomrule
  \end{tabularx}
\end{table}

\begin{table}[!tp]
  \centering
  \caption{\textbf{NeoSim benchmark: per-task success rate (\%).} Closed-loop
    success on the twelve NeoSim tasks of the companion report
    \citep{neoteaidata2026}, split into four single-arm and eight dual-arm
    tasks, for \model{} and external baselines under one aligned protocol.
    Best per task in \textbf{bold}; group and suite means in the shaded rows.}
  \label{tab:neosim}
  \small
  \setlength{\tabcolsep}{3pt}
  \renewcommand{\arraystretch}{1.15}
  \begin{tabularx}{\linewidth}{@{}l*{5}{>{\centering\arraybackslash}X}>{\columncolor{metablue!7}\centering\arraybackslash}X@{}}
    \toprule
    Task & \pizerofive{} & StarVLA-$\alpha$ & InternVLA-A1 & Xiaomi-Rob.-0 & GigaWorld-P. & \modelbf \\
    \midrule
    \multicolumn{7}{@{}l}{\textsf{\textbf{Single-arm tasks}}}\\
    Pour Ball         & 92 & 32 & 47 & 52 & 0 & \textbf{96} \\
    \begin{tabular}[c]{@{}l@{}}Unplug and\\ Plug Charger\end{tabular} & 23 & 0 & 0 & 5 & 3 & \textbf{30} \\
    Plug USB          & 74 & 72 & 36 & 62 & 34 & \textbf{81} \\
    Grasp Chip        & 86 & \textbf{93} & 12 & 49 & 71 & 88 \\
    \arrayrulecolor{black!30}\midrule\arrayrulecolor{black}
    \rowcolor{black!6}
    \textit{Single-arm mean} & 68.8 & 49.3 & 23.8 & 42.0 & 27.0 & \textbf{73.8} \\
    \midrule
    \multicolumn{7}{@{}l}{\textsf{\textbf{Dual-arm tasks}}}\\
    Insert Screw      & \textbf{26} & 8 & 0 & 0 & 0 & 24 \\
    Place Gears       & 18 & 0 & 0 & 0 & 0 & \textbf{29} \\
    Unstack Bowl      & 3 & \textbf{17} & 0 & 0 & 0 & \textbf{17} \\
    Cup Handover      & 13 & \textbf{25} & 0 & 0 & 0 & 22 \\
    Stack Cups        & 22 & 0 & 0 & 0 & 12 & \textbf{23} \\
    Stack Plates      & \textbf{96} & 0 & 0 & 0 & 0 & 94 \\
    Stack Bowls       & 93 & 0 & 8 & 42 & 0 & \textbf{96} \\
    Unstack Cup       & 3 & 31 & 0 & \textbf{71} & 9 & 10 \\
    \arrayrulecolor{black!30}\midrule\arrayrulecolor{black}
    \rowcolor{black!6}
    \textit{Dual-arm mean} & 34.3 & 10.1 & 1.0 & 14.1 & 2.6 & \textbf{39.4} \\
    \midrule
    \rowcolor{metablue!12}
    \textbf{NeoSim mean} & 45.8 & 23.2 & 8.6 & 23.4 & 10.8 & \textbf{50.8} \\
    \bottomrule
  \end{tabularx}
\end{table}

Table~\ref{tab:univtac} reports the eight UniVTAC tasks and
Table~\ref{tab:neosim} the twelve NeoSim tasks. On UniVTAC, \model{}
averages $83.1\%$, ahead of the strongest external baseline, InternVLA-A1
at $67.1\%$, with near-saturating success on several insertion tasks and
$100\%$ on grasp-and-classify. On NeoSim the twelve tasks are harder for every method; the
specialist baselines fall to between $8.6\%$ and $23.4\%$, while \model{}
holds $50.8\%$ against $45.8\%$ for \pizerofive{}. The gap between arms is
stark: on the four single-arm tasks \model{} averages $73.8\%$, but the
eight dual-arm tasks halve it to $39.4\%$ and collapse the specialists into
the single digits, InternVLA-A1 to $1.0\%$. Averaged over all twenty
tasks, \model{} reaches $63.8\%$ against $44.0\%$ for \pizerofive{}, its
strongest overall baseline. These insertion-heavy and bimanual regimes are
the contact-decided setting the latent tactile pathway targets.

\subsection{Offline Policy Improvement with \alter}
\label{sec:experiments:rl}

We evaluate \alter by
training task-specific policies from pretrained VLA and VTLA checkpoints. The
evaluation first
characterizes the pairwise progress model's response to local execution
regressions and then compares the policies on real-robot tasks.

\begin{figure}[!tp]
  \centering
  \includegraphics[width=\linewidth]{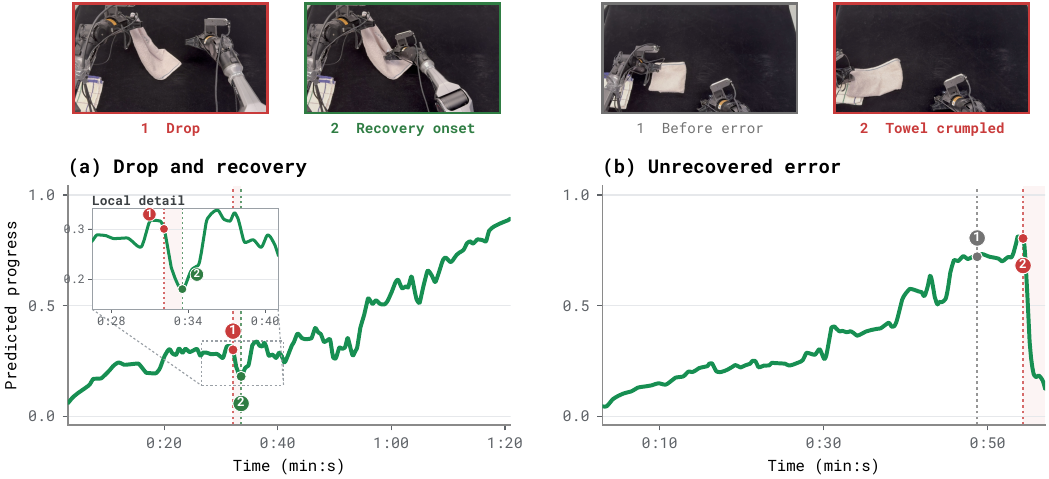}
  \caption{\textbf{Predicted progress on imperfect Towel Folding deployments.}
  Predictions from our pairwise progress model on two held-out trajectories
  containing (a)~a corrected towel drop and (b)~an unrecovered crumpling error.
  Numbered markers indicate the synchronized frames, and pale red regions
  denote degraded execution. The inset in (a) enlarges the region around
  recovery onset.}
  \label{fig:pihilcurves}
\end{figure}

\paragraph{Progress response on imperfect rollouts.}
\label{sec:policyimp:eval}
Figure~\ref{fig:pihilcurves} shows the predicted global progress on two
held-out, imperfect Towel Folding trajectories. In the
first trajectory, predicted progress drops around the towel drop, reaches a
local minimum near recovery onset, and then resumes its upward trend as the
robot recovers. In the second, predicted progress falls sharply after the
flattened towel becomes crumpled during transport and remains low through
episode termination because no recovery follows. In these examples, predicted
progress therefore decreases at both observed degradations and rises again only
when recovery follows.
Appendix~\ref{app:valueablation} provides the corresponding qualitative check
on the other two tasks.

\paragraph{Policy comparison.}
We evaluate five variants on Towel Folding, Bag Packing, and Cardboard Box Folding.
\emph{\pizerofive{}-SFT} and \emph{\model{}-SFT} are task-adapted using the
demonstration subset.
\emph{\pizerofive{}+\alter} and \emph{\model{}+\alter} apply the complete offline policy-learning procedure to the two base policies. \emph{\pizerofive{}+$\chi_0$} replaces our
progress-model supervision with the Stage Advantage supervision of \chizero{} \citep{chi0_2026}. Each offline policy-learning variant starts
from its corresponding pretrained model.

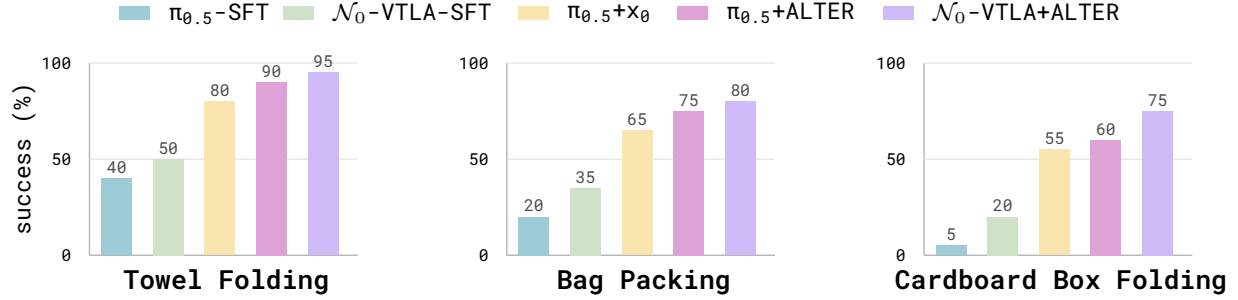
\begin{figure}[!tp]
  \centering
  \resizebox{\linewidth}{!}{\input{figures/fig_rl_policy_success}}
  \caption{\textbf{Real-robot success rates under supervised and offline policy
  learning.} All three tasks use the same $0$--$100\%$ vertical scale,
  and all five variants report measured rollouts.}
  \label{fig:pipolicy_success}
\end{figure}

Across the three tasks, \alter with a \pizerofive{} backbone obtains success rates of $90\%$, $75\%$,
and $60\%$; with \model{}, it reaches $95\%$, $80\%$, and $75\%$.
Thus, the VTLA backbone retains a consistent final advantage under the same
offline RL procedure. The \model{} policy trained with \alter completes the Towel Folding, Bag Packing,
and Cardboard Box Folding tasks at throughputs of 43, 15, and 30 successful
executions per hour, respectively. The corresponding comparison appears in
Appendix~\ref{app:valueablation}.

\subsection{Representation Analyses}
\label{sec:experiments:repr}

Our final analyses probe the learned tactile representation directly. At
the end of \stageone{} pretraining we measure three things. The first is
held-out contrastive retrieval, the accuracy form of the \stageone{}
InfoNCE objective. Over $378$ held-out queries, the predicted $z$
retrieves its matching future-tactile target $z^{*}$ of Eq.~\ref{eq:zstar}
top-1 in $92.3\%$ of cases within a pool of roughly $32$ candidates,
where chance is $3.2\%$. These are within-pool rates, not full-corpus
retrieval. The second is a control that separates anticipation from
autocorrelation. Ranking the same candidates by the current tactile
encoding $g$ alone, the predictor's own input, reaches $57\%$ top-1, and
$40\%$ in a $128$-candidate pool where the predictor holds $81\%$. The
latent therefore carries future contact information beyond its input, and
the margin grows as the pool gets harder. The third is a perturbation
probe. Swapping the tactile input moves $z$ by roughly $0.9$ in centered-cosine
distance while swapping the RGB views and prompt together moves it by no
more than about $0.2$, and the tactile-to-vision-language sensitivity
ratio settles at $4.3$ by the end of \stageone{} and measures about
$1.4$ after end-to-end joint training.

\begin{keybox}
\textbf{The predictor is grounded in touch, and it anticipates.} A latent
that merely copied its tactile input would match the current-touch
baseline at $57\%$. The measured $92.3\%$ sits far above it, and the gap
widens as the candidate pool grows. The perturbation probe argues against a
vision--language shortcut. Swapping that context barely moves $z$, and
the latent stays tactile-leaning even after joint training. The
representation entering joint training is therefore a real, anticipatory
one.
\end{keybox}

\FloatBarrier

%% file: figures/fig_rl_policy_success.tex
\begin{tikzpicture}[
  font=\rlchartfont\scriptsize,
  axis/.style={draw=black!28, line width=0.45pt},
  pi/.style={fill=rlblue, draw=none},
  vtla/.style={fill=rlgreen, draw=none},
  progress/.style={fill=rlyellow, draw=none},
  pirl/.style={fill=rlrose, draw=none},
  ours/.style={fill=rlpurple, draw=none}
]

\newcommand{\rlpizerofive}{π\textsubscript{0.5}}
\newcommand{\rlchizero}{χ\textsubscript{0}}
\newcommand{\rlmodelname}{\model{}}

\begin{scope}[xshift=0.50cm]
\node[pi, minimum width=0.32cm, minimum height=0.14cm, inner sep=0pt] at (0.40,2.88) {};
\node[anchor=west, text=black] at (0.64,2.88) {\rlpizerofive-SFT};
\node[vtla, minimum width=0.32cm, minimum height=0.14cm, inner sep=0pt] at (2.15,2.88) {};
\node[anchor=west, text=black] at (2.39,2.88) {\rlmodelname-SFT};
\node[progress, minimum width=0.32cm, minimum height=0.14cm, inner sep=0pt] at (4.65,2.88) {};
\node[anchor=west, text=black] at (4.89,2.88) {\rlpizerofive+\rlchizero};
\node[pirl, minimum width=0.32cm, minimum height=0.14cm, inner sep=0pt] at (6.35,2.88) {};
\node[anchor=west, text=black] at (6.59,2.88) {\rlpizerofive+ALTER};
\node[ours, minimum width=0.32cm, minimum height=0.14cm, inner sep=0pt] at (8.55,2.88) {};
\node[anchor=west, text=black] at (8.79,2.88) {\rlmodelname+ALTER};
\end{scope}

\newcommand{\successaxes}[3]{%
  \node[font=\rlchartfont\bfseries\footnotesize] at (1.88,0.02) {#1};
  \draw[axis] (0.42,0.30) -- (0.42,2.35);
  \draw[axis] (0.42,0.30) -- (3.28,0.30);
  \draw[black!10] (0.42,1.325) -- (3.28,1.325);
  \draw[black!10] (0.42,2.35) -- (3.28,2.35);
  \node[font=\rlchartfont\tiny, text=black, anchor=east] at (0.35,0.30) {0};
  \node[font=\rlchartfont\tiny, text=black, anchor=east] at (0.35,1.325) {#3};
  \node[font=\rlchartfont\tiny, text=black, anchor=east] at (0.35,2.35) {#2};
}

\newcommand{\scorebar}[5]{%
  \pgfmathsetmacro{\scoretop}{0.30 + 2.05*(#3)/(#5)}
  \path[#1] (#2,0.30) rectangle ({#2+0.32},\scoretop);
  \node[font=\rlchartfont\tiny, text=black!70] at ({#2+0.16},{\scoretop+0.12}) {#4};
}

\begin{scope}[xshift=0.00cm]
  \node[rotate=90, text=black] at (-0.28,1.33) {success (\%)};
  \successaxes{Towel Folding}{100}{50}
  \scorebar{pi}{0.55}{40}{40}{100}
  \scorebar{vtla}{1.10}{50}{50}{100}
  \scorebar{progress}{1.65}{80}{80}{100}
  \scorebar{pirl}{2.20}{90}{90}{100}
  \scorebar{ours}{2.75}{95}{95}{100}
\end{scope}

\begin{scope}[xshift=4.45cm]
  \successaxes{Bag Packing}{100}{50}
  \scorebar{pi}{0.55}{20}{20}{100}
  \scorebar{vtla}{1.10}{35}{35}{100}
  \scorebar{progress}{1.65}{65}{65}{100}
  \scorebar{pirl}{2.20}{75}{75}{100}
  \scorebar{ours}{2.75}{80}{80}{100}
\end{scope}

\begin{scope}[xshift=8.90cm]
  \successaxes{Cardboard Box Folding}{100}{50}
  \scorebar{pi}{0.55}{5}{5}{100}
  \scorebar{vtla}{1.10}{20}{20}{100}
  \scorebar{progress}{1.65}{55}{55}{100}
  \scorebar{pirl}{2.20}{60}{60}{100}
  \scorebar{ours}{2.75}{75}{75}{100}
\end{scope}
\end{tikzpicture}

%% file: sections/findings.tex
\section{Findings}
\label{sec:findings}

Touch changes manipulation where contact decides it. Four findings
collect the evidence.

\paragraph{Touch turns insertion from a one-shot visual commitment into
a contact-aware retry loop.} The two hardest real-robot insertions show the
clearest margins of the campaign. Socket Plugging, inserting a plug into an outlet, reaches
$85\%$ against $60\%$ for \pizerofive{}, and Board Insertion, seating an
expansion card, reaches $25\%$ on a task where ACT and \pizerofive{} both fail outright at
$0\%$. Representative rollouts reveal the behavioral difference behind
these gains. Once visual alignment suggests that insertion is possible,
\pizerofive{} commits to the downward motion. When the plug or card meets
the rim instead of entering the slot, it continues the attempt and
fails. \model{} instead responds to the unexpected contact by lifting the
arm, realigning, and attempting the insertion again. The pattern repeats
in simulation, where Insert Hole and Insert Tube reach $95\%$ and $99\%$.
The tactile pathway therefore changes insertion from a single visually
planned motion into a closed-loop process that can detect a blocked
attempt and recover from it.

\paragraph{Touch enables fine control of gripper force through aperture
adjustment.} Grip force rather than placement decides a second family of tasks. In the real-robot Bottle Standing task, the manipulated
object is an empty, compliant plastic bottle. \pizerofive{} closes the
gripper too far around the bottle mouth, pinching it firmly enough to lift
the entire bottle instead of leaving it on the supporting surface.
\model{} instead makes small, continuous adjustments to the gripper
aperture, maintaining sufficient contact to control the bottle without
gripping hard enough to lift it, and consequently stabilizes the bottle
upright. Although the tactile sensor does not measure force directly, its
deformation signal is strongly coupled to contact force and provides a
practical feedback signal for regulating grip. This behavior is consistent
with the broader results. Bottle Standing reaches $30\%$ where both
real-robot baselines remain at $0\%$, while Lift Can and Grasp Chip reach
$88\%$ in simulation.

\begin{figure}[H]
  \centering
  \includegraphics[width=\linewidth]{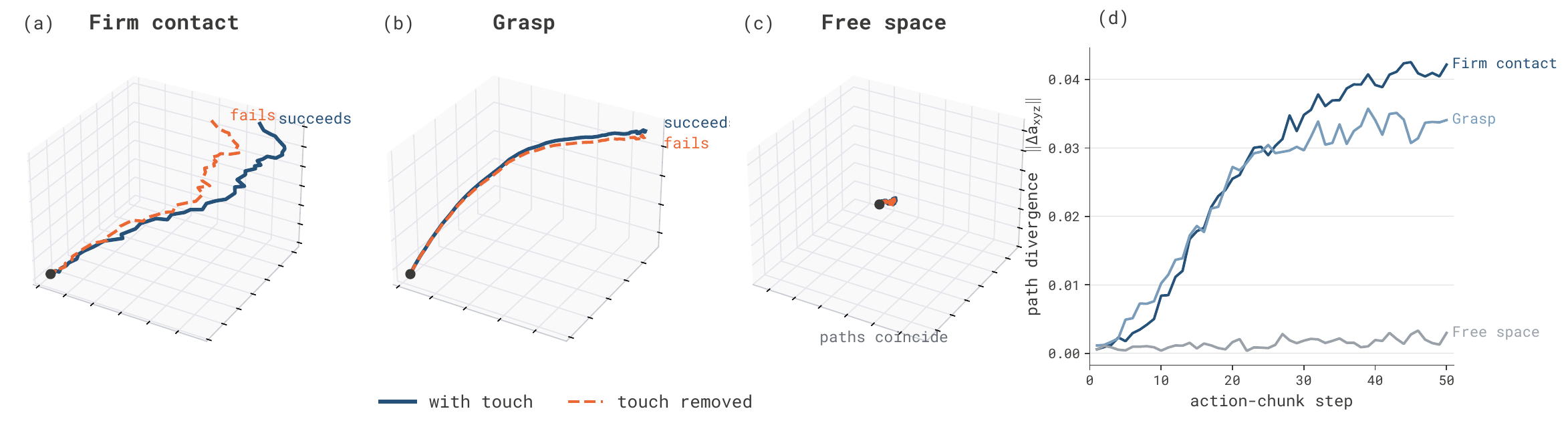}
  \caption{\textbf{A counterfactual probe of what touch changes.} At the same
    observation and under identical sampling noise, the policy's predicted
    end-effector path with touch is compared with its touch-removed
    counterfactual at three moments of one episode. At firm contact and
    grasp the paths separate, and the with-touch path is the one that
    succeeds. In free space, drawn at the same spatial scale as (a), the
    two coincide. Panel (d) plots the per-step divergence of the two
    paths over the 50-step chunk: it grows through the contact-critical
    chunks and stays near zero in free space. Axes are normalized action
    units.}
  \label{fig:counterfactual}
\end{figure}

\paragraph{Touch changes the action exactly when contact decides the
outcome.} Figure~\ref{fig:counterfactual} probes the mechanism directly. At
the same observation and under identical sampling noise, the predicted
action path with touch diverges from its touch-removed counterpart at the contact-critical
moments of firm contact and grasp, and coincides with it in free space. Because everything else is held fixed,
the comparison isolates the tactile pathway's contribution to the sampled
action. In the probed episode it is silent in free space and decisive at
contact.

\paragraph{Stronger tactile pretraining remains advantageous under \alter.}

Under clean-demonstration SFT, \model{} leads \pizerofive{} by $10$, $15$, and $15$ points on Towel Folding, Bag Packing, and Cardboard Box Folding.
When each backbone is trained with \alter from its corresponding pretrained checkpoint, \model{} reaches $95\%$, $80\%$, and $75\%$, compared with
$90\%$, $75\%$, and $60\%$ for \pizerofive{}. The advantage associated with the stronger VTLA checkpoint therefore persists under the same offline RL procedure.

%% file: sections/related.tex
\section{Related Work}
\label{sec:related}

\paragraph{Vision--language--action policies.}
Modern vision--language--action (VLA) policies build on action-chunk and
generative visuomotor architectures established by ACT \citep{act2023} and
Diffusion Policy \citep{diffusionpolicy2023}. ACT predicts temporally coherent
action chunks, whereas Diffusion Policy models multimodal action sequences
through iterative denoising. RT-1 \citep{rt1} established transformer
policies for real-world control at scale, and RT-2 \citep{rt2_2023}
subsequently connected internet-pretrained vision--language representations
to robot control, while RT-X \citep{openx2023} and Octo \citep{octo2024}
scaled generalist policies across heterogeneous robot datasets. Subsequent VLA systems have diversified
along two broad axes: their action-generation mechanisms and their strategies
for scaling, transfer, and deployment. OpenVLA \citep{openvla2024} provides an
open autoregressive backbone, while \pinought{} \citep{pi0_2024} and
\pizerofive{} \citep{pi05_2025} generate continuous action chunks through flow
matching. Autoregressive action
tokens reuse the language-model interface and scale naturally with pretrained
backbones, whereas continuous generative heads model low-level commands directly
in continuous space and can capture multiple distinct, valid action sequences.
Beyond action generation,
UniVLA \citep{univla2025} and Qwen-VLA \citep{qwenvla2026} extend transfer
across tasks, environments, and embodiments. StarVLA \citep{starvla2026}
emphasizes modularity, MiMo-Embodied \citep{mimoembodied2025} targets
cross-embodiment scaling, and Pragmatic VLA \citep{pragmaticvla2026} focuses on
practical deployment. Recent foundation-scale systems push data and model
size further: Xiaomi-Robotics-1 \citep{xiaomi2026robotics1} pretrains on more
than $100{,}000$ hours of real-world trajectories, its U0 variant
\citep{li2026xiaomiu0} couples action synthesis with a world foundation model,
CronusVLA \citep{cronusvla} aggregates multi-frame context, InternVLA-A1.5
\citep{internvla_a15} unifies understanding, latent foresight, and action, and
GR-3 \citep{gr3_2025}, GR00T~N1 \citep{groot_n1_2025}, and SmolVLA
\citep{smolvla2025} span large generalist to affordable efficient policies.
Benchmark suites such as VLABench \citep{vlabench2024} and LIBERO-Plus
\citep{liberoplus2025} track this progress, probing long-horizon reasoning
and robustness under controlled perturbations.
Across these systems,
generalization is pursued either by expanding heterogeneous pretraining
mixtures or by learning task and embodiment abstractions that can be reused
during adaptation. Together, these lines combine semantic grounding,
cross-robot data scaling, and generative continuous control. Their observation
spaces, however, remain centered on cameras, language, and robot state, leaving
contact and slip only indirectly observable. \model{} retains the
language-conditioned, flow-matching \pizerofive{} backbone but adds a dedicated
high-resolution tactile pathway.

\paragraph{Tactile-conditioned and predictive VTLA policies.}
Work toward tactile-conditioned vision--language--action control combines
transferable tactile representations and scalable data collection with policies
that either react to measured touch or predict how contact will evolve.
Touch and Go \citep{touchandgo2022} and Visuo-Tactile Transformers
\citep{vtt2022} learn joint visual--tactile features. VITaL
\citep{vital2024}, Sparsh \citep{sparsh2024}, AnyTouch \citep{anytouch2025},
and UniTouch \citep{unitouch2024} emphasize transfer across tasks or sensors,
and FTP-1 \citep{ftp1_2026} scales cross-sensor transfer to a generalist
tactile policy pretrained on thousands of hours of tactile data.
FreeTacMan \citep{freetacman2025}, exUMI \citep{exumi2025}, ViTaMIn
\citep{vitamin2025}, Touch in the Wild \citep{touchinthewild2025}, and UniVTAC
\citep{univtac2026} use robot-free collection or simulation to reduce
contact-rich data costs. Reactive Diffusion Policy \citep{rdp2025} uses touch
for fast feedback, 3D-ViTac \citep{vitac3d2024} performs spatial fusion,
TouchGuide \citep{touchguide2026} provides inference-time steering, and
Multi-Modal Policy Consensus \citep{policyconsensus2025} fuses controllers,
and T-Rex \citep{trex2026} pairs a variable-rate architecture with
high-frequency touch for tactile-reactive dexterous control.
TLA \citep{tla2025}, Tactile-VLA \citep{tactilevla2025}, VLA-Touch
\citep{vlatouch2025}, and OmniVTLA \citep{omnivtla2025} further integrate
observed touch into language-conditioned control, while Neural Feels
\citep{neuralfeels2024} targets fine-grained state estimation. Seeing Touch
from Motion \citep{seeingtouch2026} reads fine-grained contact states from
tactile motion correlation, fusing touch and vision in a modality-aware
mixture-of-transformers policy. Force-aware
variants condition on measured or estimated force: ForceVLA \citep{forcevla}
adds a force-aware mixture-of-experts, and TA-VLA \citep{tavla} a torque-aware
interface for contact-rich control. Together, they
establish the value of current touch for closed-loop correction and grounding.
Predictive methods complement this reactive signal by estimating contact before
it unfolds: Imagine2Touch \citep{imagine2touch2024} predicts local tactile
readings ahead of contact, and TTP \citep{ttp2026} transfers a future-tactile
expert to robot policies. HapticVLA \citep{hapticvla2026} likewise estimates
rather than measures touch, though its target is the current signal: a tactile
token inferred from vision and robot state replaces the sensor at deployment. Visuo-Tactile World Models
\citep{vtwm2026}, OmniVTA \citep{omnivta2026}, Tactile-WAM
\citep{tactilewam2026}, Dream-Tac \citep{dreamtac2026}, VT-WAM
\citep{vtwam2026}, TacForeSight \citep{tacforesight}, and ContactWorld
\citep{contactworld} instead forecast richer dynamics, sensory futures, and
actions. A parallel predictive line forecasts future observations rather than
latents: video-generation policies such as UniPi \citep{unipi}, GR-1
\citep{gr1}, and GR-2 \citep{gr2} synthesize pixel futures to guide action, and
Unified World Models \citep{uwm} couple video and action diffusion, and
World Guidance \citep{WoG} models the world in a learned
condition space rather than in pixels to guide action generation. Against this backdrop, \model{} occupies a lighter point in the design space. Rather
than training a tactile-specific encoder or decoding a full sensory trajectory,
it repurposes frozen DINOv2 features \citep{dinov2_2023}, retains measured
touch, and uses vision and language to predict its net change over the action
horizon. The resulting compact latent conditions the action expert directly.
This latent-target formulation is related to I-JEPA \citep{ijepa2023}, V-JEPA
\citep{vjepa2024}, V-JEPA 2 \citep{vjepa2_2025}, LeJEPA
\citep{lejepa2025}, and LeWorldModel \citep{leworldmodel2026}. Its distinction
lies in the target, placement, training recipe, and scale within a shared
language-conditioned VLA.

\paragraph{Offline RL and VLA policy improvement from deployment data.}
Deployment exposes distribution shifts and long-tail failures absent from
static demonstrations, while human interventions provide corrective data.
RECAP \citep{pistar06_2025} learns from demonstrations, autonomous experience,
and interventions through advantage-conditioned policy extraction. SOP
\citep{sop2026} scales collection
and learning across robot fleets. LWD \citep{lwd2026} uses value-guided updates
for flow-based VLAs. VLA-RL \citep{vlarl2025} performs trajectory-level RL
with segment-derived process rewards. A complementary line converts sparse
outcomes into dense process supervision. Vision--language models can serve as
in-context value estimators \citep{gvl2025}, while Robo-Dopamine
\citep{robodopamine2025} learns step-aware, multi-view rewards. SARM
\citep{sarm2026} models stages and within-stage progress for filtering and
reweighting variable-duration demonstrations. ARM \citep{arm2026} labels
relative progression, regression, or stagnation. Learned progress or value then
filters
demonstrations in GR-RL \citep{grrl2025}, weights flow matching in ProgVLA
\citep{progvla2026}, guides actions in ProgressVLA \citep{progressvla2026}, and
supports offline-to-online adaptation in Robo-ValueRL \citep{robovalue2026}.
Closest to our setting, \chizero{} \citep{chi0_2026} predicts stage-aware relative progress from paired observations, while RECAP \citep{pistar06_2025} conditions VLA training on binarized advantages. \chizero{} derives its progress targets by spacing stages uniformly and using elapsed time. \alter instead weights stages by demonstrated duration and adds local comparison supervision around tactile-detected object-drop events and logged HIL corrections for advantage-conditioned offline policy learning.

%% file: sections/conclusion.tex
\section{Conclusion}
\label{sec:conclusion}

We presented \model{}, a vision--tactile--language--action foundation model
that predicts outcomes instead of reacting to them. For perception,
\emph{latent tactile
tokens} predict contact. A frozen visual backbone
with a trainable projection turns the gripper's contact-difference images from our
self-developed visuo-tactile sensor into tokens, and a lightweight predictor
compresses these
tokens, together with the vision--language context, into a latent $z$ that
conditions a flow-matching action expert directly, never entering the
vision--language prefix, so that touch is treated as a prediction target
rather than as additional observation context. For policy improvement, \alter
formulates learning from deployment corpus as advantage-conditioned offline RL. Its pairwise progress model estimates global task phase and local
execution change, which are converted into stage-relative advantage conditions
for offline policy learning. A three-stage recipe brings
this newly initialized pathway online without destabilizing the pretrained
\pizerofive{} backbone. \stageone{} grounds the predictor against a
future-tactile target, \stagetwo{} freezes the tactile perception stack and
masks the vision--language pathway to align the resulting latents with the
action expert, and \stagethree{}
trains the full policy jointly. The
recipe rests on a canonical cross-embodiment action space
and NeoData, a large-scale curated multi-platform corpus.
Representation-level evidence
already supports the design. After \stageone{}, the latent tokens retrieve
their matching future-tactile target at $92.3\%$ top-1 accuracy, far above
the $3.2\%$ chance level.
The closed-loop results point the same way. \model{} attains the
highest average on the simulation suite by a margin of more
than nineteen points over the strongest baseline, and goes unbeaten
across the nine NeoReal real-robot tasks.

\paragraph{Future work.} Two directions follow directly from this report.
First, the free-latent and supervised variants defined here are two points in
a broader design space for tactile representation learning, and we see the
predictive-latent framing, conditioning action generation on a compact
estimate of the net contact change over the chunk horizon rather than on raw
sensory tokens, as a direction worth pursuing beyond the specific predictor
architecture used in this report.
Second, \alter can be extended to a broader range of tasks
to study its effectiveness across different manipulation settings.

%% file: sections/contributors.tex
\section*{Contributors}

\textbf{Pretraining.} Heng Zhou, Silong Dai, Yutao Fan, Yiran Qin, Shunlin Lu. \\
\textbf{Offline RL.} Yutao Fan, Longjie Su, Heng Zhou, Yiming Wu, Yiran Qin, Shunlin Lu.\\
\textbf{Post-Training (Simulation).} Heng Zhou, Bruno N.Y. Chen, Zhemeng Zhang, Yiran Qin, Shunlin Lu.\\
\textbf{Post-Training (Real Robot).} Jiongwei Lu, Silong Dai, Heng Zhou, Yutao Fan,  Zhemeng Zhang, Shengqi Xu, Boyu Mi, Bruno N.Y. Chen, Li Kang, Yanjun Li.\\
\textbf{Data processing.} Yutao Fan, Heng Zhou, Rui Li, Xiufeng Song, Tianyu Yang, Wenjie Zhou,  Yifan Wang, Yiming Wu, Xin Wang, Bruno N.Y. Chen.\\
\textbf{Academic Supervision.} Ziyi Ye, Guoxiang Dong, Xiaosong Jia, Wenming Chen. \\
\textbf{Project Lead.}  Yiran Qin, Shunlin Lu, Shihao Zhao, Daoguo Dong, Zuxuan Wu.

%% file: sections/appendix.tex
\section{Data Card}
\label{app:datacard}

Table~\ref{tab:datacard} summarizes NeoData, the released corpus, at the
level of detail expected of a data card. Collection protocols, sensor
specifications, and per-repository provenance are documented in full in the
companion data technical report \citep{neoteaidata2026}.

\begin{table}[H]
  \centering
  \caption{Data card for NeoData, the released corpus.}
  \label{tab:datacard}
  \begin{tabular}{p{3.2cm}p{10.2cm}}
    \toprule
    Field & Value \\
    \midrule
    Modalities & 3 RGB views per episode (one third-person, two
      wrist-mounted), a language instruction, robot proprioceptive state,
      and 2 tactile streams per episode on single-arm platforms or 4 on
      dual-arm platforms, each from the self-developed visuo-tactile
      sensor (Section~\ref{sec:data:tactile}). \\
    Format & LeRobot dataset format \citep{lerobot2024}, holding per-view video
      files, tabular action/state records, and per-episode metadata. \\
    Frame rate & $30$\,fps, uniform across every modality, view, and
      platform. \\
    Resolution & Raw capture up to $640\times480$ per RGB view. Tactile
      streams are captured at native sensor resolution rather than padded
      or upsampled at collection time. All streams are resized to
      $224\times224$ for model consumption. \\
    Platforms & Single- and dual-arm configurations spanning the ARX X5,
      UR5e, Flexiv, Franka, and Piper arms and the Neo TacUMI handheld
      gripper (see the companion data report \citep{neoteaidata2026}). \\
    Scale & Composition and
      episode accounting in the companion data report
      \citep{neoteaidata2026}. \\
    Action/state schema & Canonical $32$-dimensional container. The first
      $20$ dimensions hold one $10$-dimensional slot per arm, each carrying a
      $3$-dimensional position, a $6$-dimensional rot6d rotation, and a
      $1$-dimensional gripper channel, and the remaining $12$ are always zero
      (Section~\ref{sec:data:canonical}). \\
    Language annotations & Curated per-task English instructions, with
      several paraphrase variants per task assigned deterministically per
      episode. \\
    \bottomrule
  \end{tabular}
\end{table}

\section{Compute Infrastructure}
\label{app:compute}

\model{} is trained with multi-node distributed training at cluster scale on
modern accelerators. Simulation post-training and smaller fine-tuning jobs run
at correspondingly smaller scale on the same class of hardware. We describe
the infrastructure at this level of generality and omit hardware models,
device counts, memory footprints, and throughput figures.

\section{Data-Engineering Pitfalls}
\label{app:pitfalls}

Section~\ref{sec:data:gates} defers to this appendix the individual invariants
behind data quality verification. Every converted repository is checked
against the invariants in Table~\ref{tab:pitfalls} before it is allowed into
training. Each pairs a red line with the symptom it produces when violated.

\begin{table}[H]
  \centering
  \caption{Data-engineering invariants enforced before training and the
    symptom each produces when violated.}
  \label{tab:pitfalls}
  \small
  \begin{tabular}{p{6.0cm}p{7.0cm}}
    \toprule
    Red line & Symptom if violated \\
    \midrule
    Actions stored as absolute end-effector poses on disk & Training loss
      converges normally, but the resulting policy is unusable on
      hardware. \\
    A single coordinate frame and unit convention (meters) for state and
      action & The normalization sanity check on the mean position
      channel fails, well above its expected near-zero value. \\
    Maximum group-of-pictures size of $30$ frames per video & Training
      throughput collapses under periodic stalls that starve the trainer of
      data, since every random read decodes from the nearest keyframe. \\
    Tactile frames stored at native sensor resolution, never padded or
      upsampled at collection & Wasted resolution, or, if padding is
      applied, spurious fixed content baked into every frame. \\
    Every episode begins with a static, contact-free baseline period &
      The zero-contact reference frame used throughout the
      contact-difference tactile processing of
      Section~\ref{sec:data:tactile} is no longer valid. \\
    Corrupt video files are culled before entering training & A single
      corrupt file can silently kill a data-loading worker mid-run. \\
    Any in-place data change is followed by purging the cached,
      pre-built training dataset & Training silently reads a stale
      cached dataset. The loss trajectory starts anomalously high and
      plateaus without converging. \\
    Normalization statistics are computed under the same chunk-relative
      action transform used at training time & Training proceeds and the
      loss looks healthy, but the model learns at the wrong action scale. \\
    Dataset output root matches the configured dataset root directory rather
      than a nested subdirectory & Repositories are written to a nested
      path and become undiscoverable by the training data loader. \\
    \bottomrule
  \end{tabular}
\end{table}

\section{Deployment Protocol}
\label{app:deploy}

Section~\ref{sec:training} defers deployment-time details of the trained
policy to this appendix, namely the execution contract for a predicted action
chunk, the tactile baseline convention used at serve time, and the general
shape of the serving interface.

\paragraph{Full action-chunk execution.} At serve time the action expert
emits an action chunk spanning the full training horizon of $H=50$ steps
set in Section~\ref{sec:method:base}, from one prefix encoding and one denoising
solve of ten Euler steps per policy request. The deployment
contract requires the controller to execute every step of a returned chunk
before requesting the next one. This is not a convenience default. Executing
only a truncated prefix of each chunk, fewer than the full $50$ steps, before
re-predicting can leave the arm at the target position with the gripper never
closed, because the gripper-closing commands are concentrated in the tail of
the chunk that early re-planning discards. This behavior is a property of the
execution schedule, not of the underlying model, which closes the gripper
correctly under full-chunk execution.

\paragraph{Tactile baseline reset.} The tactile pathway of
Section~\ref{sec:method:tactile} conditions on a per-episode
contact-difference baseline rather than on raw tactile frames. At serve
time, a reset issued at the start of an episode clears any previously
stored baseline. The first observation received after a reset is captured
as the new zero-contact reference for every active tactile view, so the
tactile difference at that instant is exactly zero, matching the convention
established at collection time in Section~\ref{sec:data:tactile}. Every
subsequent observation in the episode is differenced against that same
stored baseline until the next reset. If a given view's tactile stream is
unavailable, the controller falls back to treating that view as its own
baseline, that is, no contact, rather than failing the request.

\paragraph{Serving interface.} The trained policy is served behind a
message-based request/response interface. A reset message clears the
stored tactile baseline for a new episode. A predict request carries the
current observation, comprising camera images, active tactile views,
robot state, and an optional language instruction, and the interface
returns the predicted action chunk together with basic timing
information. We describe the
interface at this level of generality and omit transport-level and
deployment-script details.

\section{Per-Task Results and Scoring Rubrics}
\label{app:pertask}

This appendix lists the point-based progress rubrics that define the progress
score of Section~\ref{sec:experiments:protocol} for the rubric-scored NeoReal
tasks. Each rubric decomposes a task into a fixed sequence of
subtask checkpoints with a point value each, so that the progress score is a
reproducible sum of checkpoint credit rather than a subjective judgment.

\paragraph{Rubric-scored NeoReal tasks.}
Every NeoReal task is scored on a $100$-point stage rubric.
Table~\ref{tab:rubrics} lists the checkpoint sequences for two
representative long-horizon tasks. The per-checkpoint point weights, which
distribute the $100$ points within each task, are documented in the
companion data report \citep{neoteaidata2026}.

\begin{table}[H]
  \centering
  \caption{Checkpoint sequences for two representative long-horizon NeoReal
    task rubrics. Each task's stage rubric distributes $100$ points across
    the checkpoints listed.}
  \label{tab:rubrics}
  \begin{tabular}{ll}
    \toprule
    Task & Checkpoint sequence \\
    \midrule
    \multirow{5}{*}{Towel Folding}
      & Take one towel from the pile \\
      & Lift it \\
      & Shake it flat \\
      & Fold it into a square \\
      & Place it aside \\
    \midrule
    \multirow{5}{*}{Bag Packing}
      & Pull the bag close \\
      & Open it \\
      & Put items in \\
      & Arrange them \\
      & Place the bag aside \\
    \bottomrule
  \end{tabular}
\end{table}

\begin{figure}[!tp]
  \centering
  \includegraphics[width=\linewidth, trim=12 80 5 46, clip]{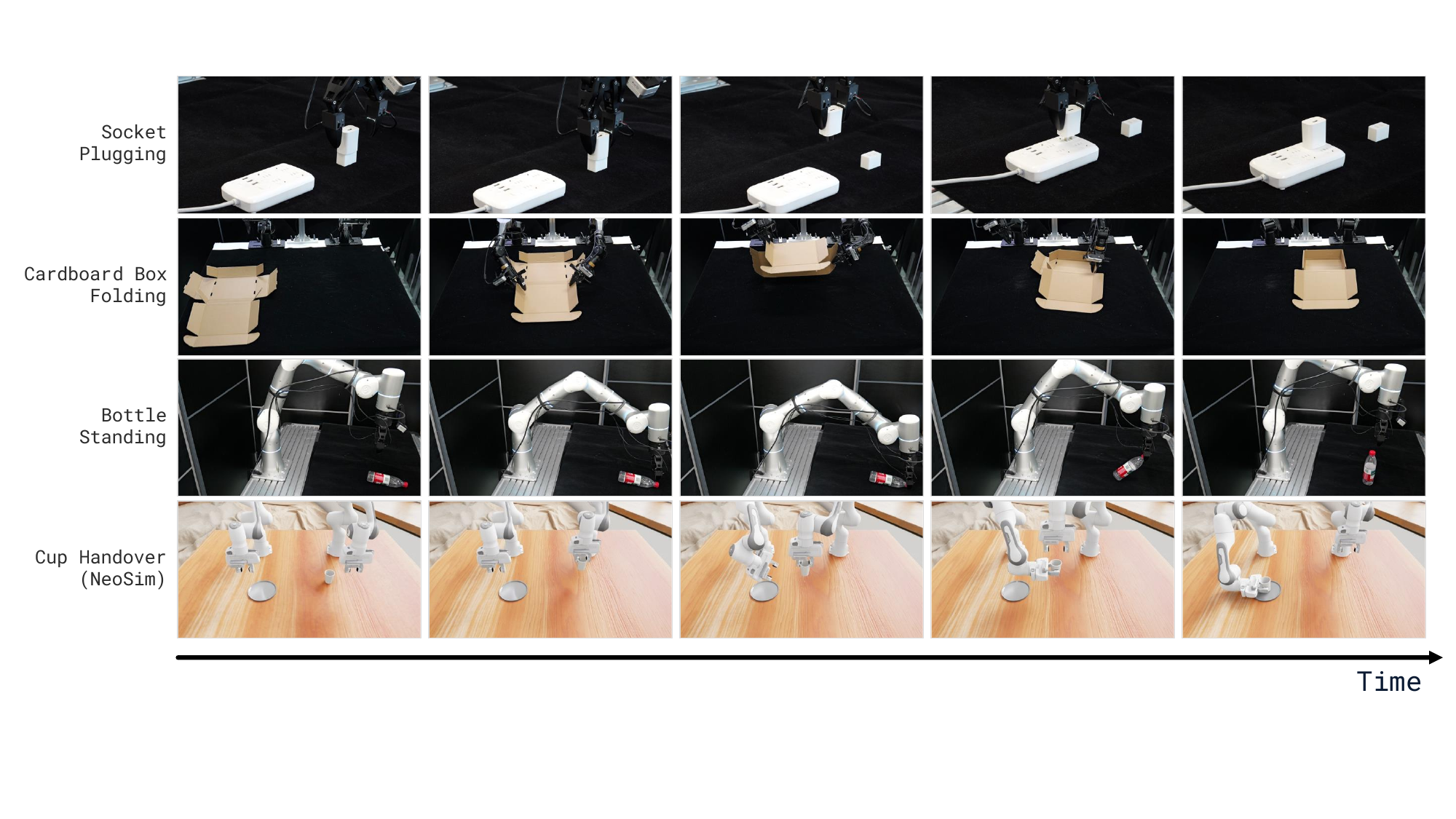}
  \caption{\textbf{Representative rollouts.} Keyframe strips
    of \model{} executions over time: Socket Plugging,
    Cardboard Box Folding, and Bottle Standing from NeoReal, and the
    dual-arm Cup Handover task in NeoSim.
    Aggregate results are plotted in Figure~\ref{fig:neoreal}.}
  \label{fig:rollouts}
\end{figure}

\paragraph{Representative rollouts.}
Figure~\ref{fig:rollouts}, referenced from
Section~\ref{sec:experiments:neoreal}, complements the aggregate scores
with keyframe strips of \model{} executions on three NeoReal tasks and one
NeoSim task.

\section{\alter Offline Policy-Learning Details}
\label{app:valueablation}

\paragraph{Deployment-corpus composition.}
The screened deployment corpus contains the three mutually exclusive
episode categories of Section~\ref{sec:policyimp:data}. Towel Folding uses 351 clean
demonstrations, 286 autonomous rollouts, and 180 HIL rollouts. Bag Packing uses
213, 243, and 238 episodes, respectively. Cardboard Box Folding uses 550,
587, and 397 episodes, respectively. The HIL totals include the separately
collected staged-recovery episodes that begin from selected error states;
these subsets contain 80, 72, and 110 episodes for the three tasks,
respectively.

\begin{figure}[H]
  \centering
  \resizebox{\linewidth}{!}{\input{figures/fig_rl_policy_throughput}}
  \caption{\textbf{Successful-execution throughput under supervised and offline
  policy learning.} All bars report measured successful task completions per
  hour. Vertical-axis ranges differ across tasks and are labeled separately.}
  \label{fig:pipolicy_throughput}
\end{figure}
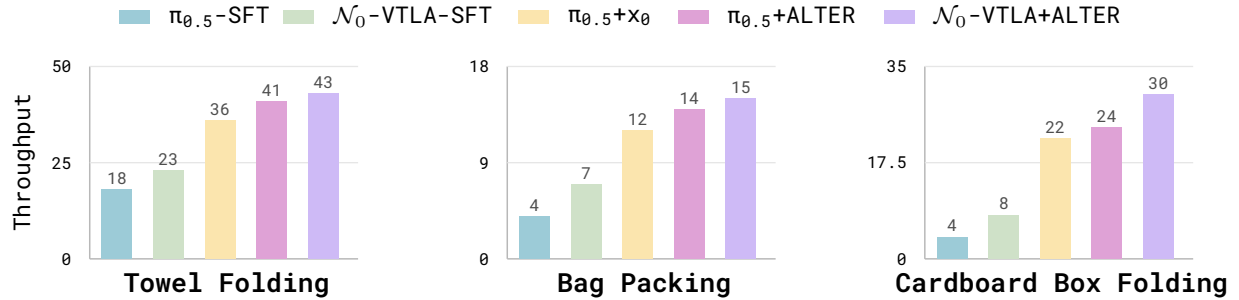

\paragraph{Successful-execution throughput.}
Figure~\ref{fig:pipolicy_throughput} complements the main-text success-rate
comparison with the measured number of successful task completions per hour
under the evaluation protocol.

\begin{figure}[H]
  \centering
  \includegraphics[width=\linewidth]{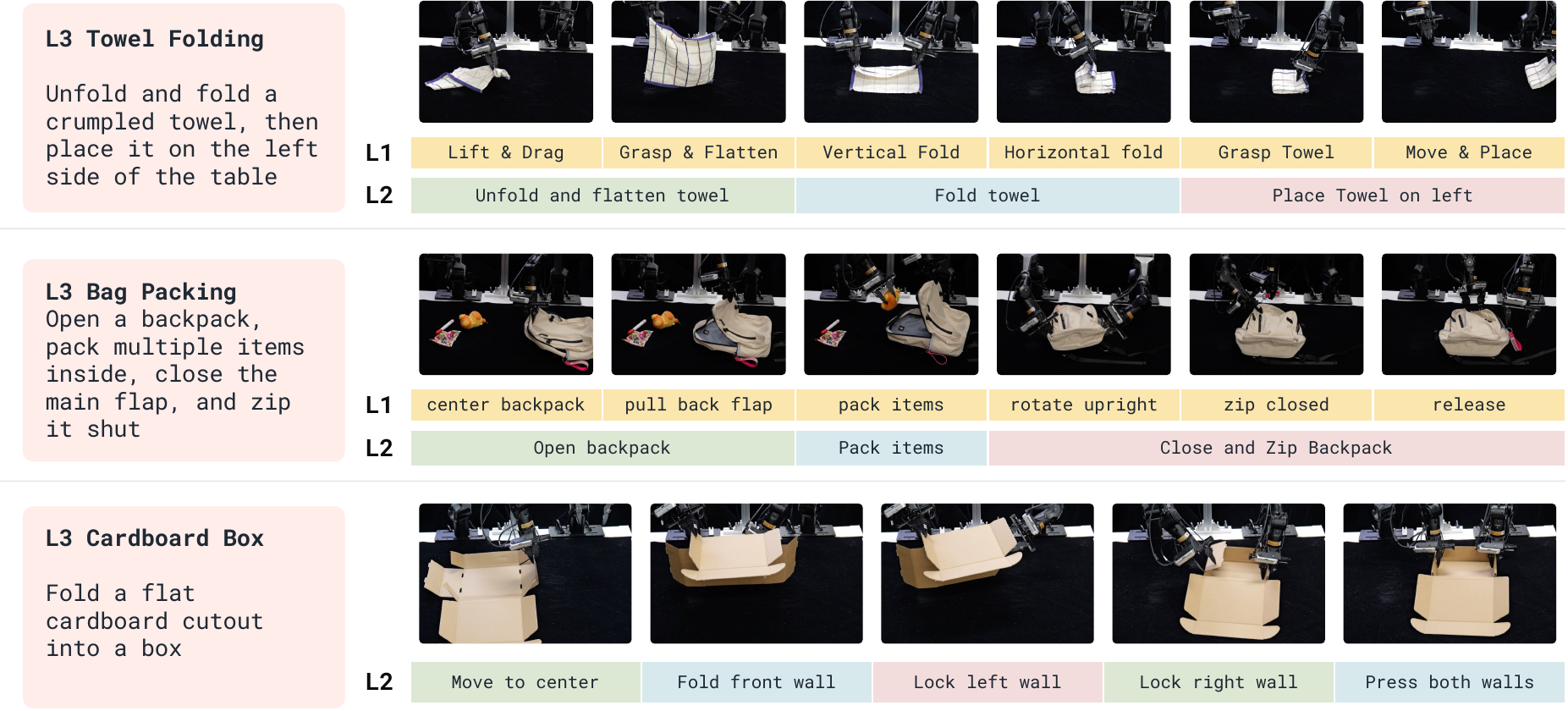}
  \caption{\textbf{Representative stage annotations used for progress-model
  supervision.} Towel Folding and Bag Packing use L1 action stages, with their
  parent L2 stages shown below. Cardboard Box Folding uses L2 stages.}
  \label{fig:piannotations}
\end{figure}

\paragraph{Representative stage annotations.}
Figure~\ref{fig:piannotations} shows the task-specific stage granularity used
to construct the clean-demonstration progress targets of
Section~\ref{sec:policyimp}. Towel Folding and Bag Packing use L1 stages.
Cardboard Box Folding uses L2 stages because its L1 decomposition is overly
fine-grained and its L2 stages already provide sufficient task structure.

\paragraph{Implementation details.}
For event supervision we use margin $m=0.02$. In implementation, event pairs
comprise approximately $5\%$ of sampled pairs and receive a per-sample loss
multiplier of $0.1$. Before assigning stages, we apply a five-frame median
filter to the global-progress predictions. We then enforce non-decreasing stage
indices by replacing each predicted index with the maximum index observed up to
that frame. These operations prevent short-term prediction noise from moving a
sample back to an earlier stage. They do not modify the local-change estimate
used for within-stage ranking. Because a staged-recovery episode begins from a
selected error state, its initial global progress cannot be assumed to be zero.
We estimate the offset $\delta_e$ in
Equation~\ref{eq:pi-advantage-condition} by comparing its first observation
with the starts of three clean reference demonstrations. If fewer
than $H$ future frames remain, the local score is normalized by the ratio of
the full action-chunk horizon to the available horizon. The terminal
self-comparison is zero. During policy training, the advantage tag is omitted
with probability $0.3$, so the policy also sees the original task prompt
without the additional text input.

\begin{figure}[H]
  \centering
  \includegraphics[width=\linewidth]{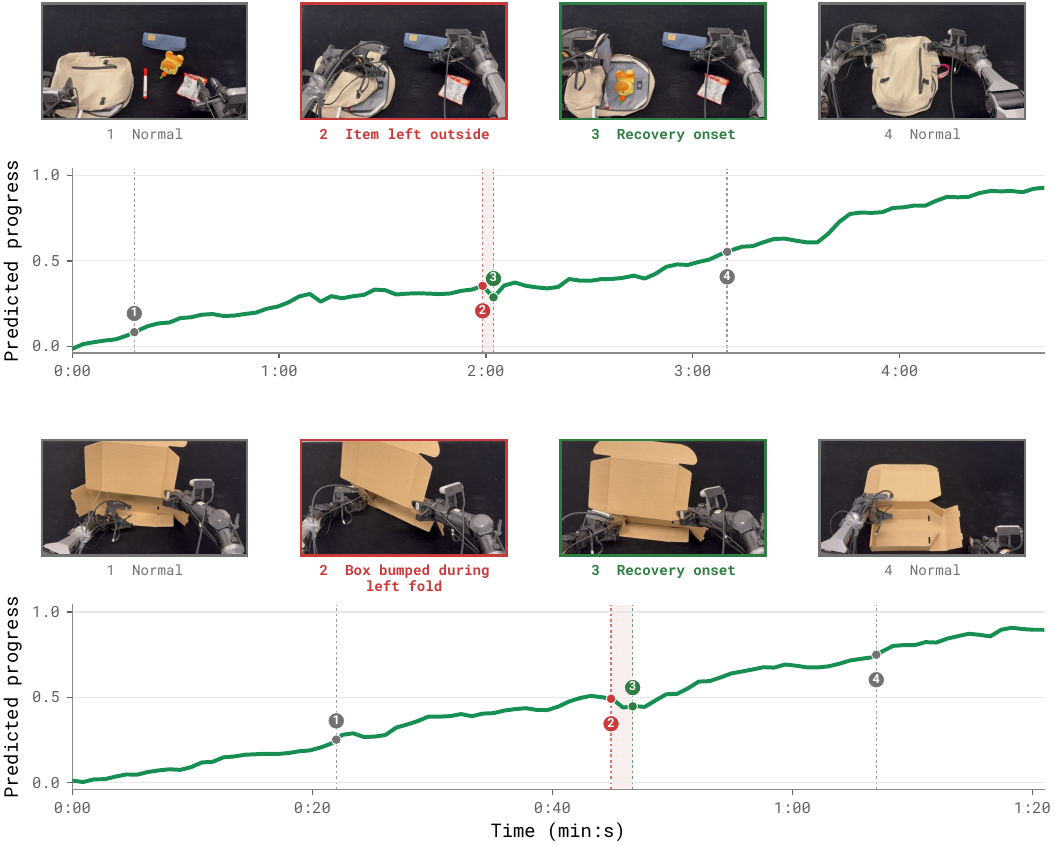}
  \caption{\textbf{Predicted progress on additional deployment tasks.}
  One evaluation episode per task, held out from progress-model training and
  scored by our pairwise progress model. The curve is the
  predicted global progress $\hat\phi_t$ against the episode's first frame.
  Numbered curve markers correspond to the synchronized frames above each
  panel. Gray denotes normal execution, red the onset of
  degradation, and green the onset of recovery. The pale red interval spans
  the detected degradation. In the Bag Packing episode, the policy starts
  closing the bag while an item remains on the table before recovering. In
  Cardboard Box Folding, the arm accidentally knocks the box out of alignment
  while folding the left side, and the subsequent correction restores its
  pose.}
  \label{fig:piourscurvesappendix}
\end{figure}

\paragraph{Progress trajectories on additional tasks.}
Figure~\ref{fig:piourscurvesappendix} shows our pairwise progress model on the
two tasks omitted from the main-text diagnostic. In both cases, predicted
progress decreases when execution degrades and rises again after the robot
returns to a productive execution state.

%% file: figures/fig_rl_policy_throughput.tex
\begin{tikzpicture}[
  font=\rlchartfont\scriptsize,
  axis/.style={draw=black!28, line width=0.45pt},
  pi/.style={fill=rlblue, draw=none},
  vtla/.style={fill=rlgreen, draw=none},
  progress/.style={fill=rlyellow, draw=none},
  pirl/.style={fill=rlrose, draw=none},
  ours/.style={fill=rlpurple, draw=none}
]

\newcommand{\rlpizerofivethroughput}{π\textsubscript{0.5}}
\newcommand{\rlchizerothroughput}{χ\textsubscript{0}}
\newcommand{\rlmodelnamethroughput}{\model{}}

\begin{scope}[xshift=0.50cm]
\node[pi, minimum width=0.32cm, minimum height=0.14cm, inner sep=0pt] at (0.40,2.88) {};
\node[anchor=west, text=black] at (0.64,2.88) {\rlpizerofivethroughput-SFT};
\node[vtla, minimum width=0.32cm, minimum height=0.14cm, inner sep=0pt] at (2.15,2.88) {};
\node[anchor=west, text=black] at (2.39,2.88) {\rlmodelnamethroughput-SFT};
\node[progress, minimum width=0.32cm, minimum height=0.14cm, inner sep=0pt] at (4.65,2.88) {};
\node[anchor=west, text=black] at (4.89,2.88) {\rlpizerofivethroughput+\rlchizerothroughput};
\node[pirl, minimum width=0.32cm, minimum height=0.14cm, inner sep=0pt] at (6.35,2.88) {};
\node[anchor=west, text=black] at (6.59,2.88) {\rlpizerofivethroughput+ALTER};
\node[ours, minimum width=0.32cm, minimum height=0.14cm, inner sep=0pt] at (8.55,2.88) {};
\node[anchor=west, text=black] at (8.79,2.88) {\rlmodelnamethroughput+ALTER};
\end{scope}

\newcommand{\throughputaxes}[3]{%
  \node[font=\rlchartfont\bfseries\footnotesize] at (1.88,0.02) {#1};
  \draw[axis] (0.42,0.30) -- (0.42,2.35);
  \draw[axis] (0.42,0.30) -- (3.28,0.30);
  \draw[black!10] (0.42,1.325) -- (3.28,1.325);
  \draw[black!10] (0.42,2.35) -- (3.28,2.35);
  \node[font=\rlchartfont\tiny, text=black, anchor=east] at (0.35,0.30) {0};
  \node[font=\rlchartfont\tiny, text=black, anchor=east] at (0.35,1.325) {#3};
  \node[font=\rlchartfont\tiny, text=black, anchor=east] at (0.35,2.35) {#2};
}

\newcommand{\ratebar}[5]{%
  \pgfmathsetmacro{\ratetop}{0.30 + 2.05*(#3)/(#5)}
  \path[#1] (#2,0.30) rectangle ({#2+0.32},\ratetop);
  \node[font=\rlchartfont\tiny, text=black!70] at ({#2+0.16},{\ratetop+0.12}) {#4};
}

\begin{scope}[xshift=0.00cm]
  \node[rotate=90, text=black] at (-0.28,1.33) {Throughput};
  \throughputaxes{Towel Folding}{50}{25}
  \ratebar{pi}{0.55}{18}{18}{50}
  \ratebar{vtla}{1.10}{23}{23}{50}
  \ratebar{progress}{1.65}{36}{36}{50}
  \ratebar{pirl}{2.20}{41}{41}{50}
  \ratebar{ours}{2.75}{43}{43}{50}
\end{scope}

\begin{scope}[xshift=4.45cm]
  \throughputaxes{Bag Packing}{18}{9}
  \ratebar{pi}{0.55}{4}{4}{18}
  \ratebar{vtla}{1.10}{7}{7}{18}
  \ratebar{progress}{1.65}{12}{12}{18}
  \ratebar{pirl}{2.20}{14}{14}{18}
  \ratebar{ours}{2.75}{15}{15}{18}
\end{scope}

\begin{scope}[xshift=8.90cm]
  \throughputaxes{Cardboard Box Folding}{35}{17.5}
  \ratebar{pi}{0.55}{4}{4}{35}
  \ratebar{vtla}{1.10}{8}{8}{35}
  \ratebar{progress}{1.65}{22}{22}{35}
  \ratebar{pirl}{2.20}{24}{24}{35}
  \ratebar{ours}{2.75}{30}{30}{35}
\end{scope}
\end{tikzpicture}

%% file: main.bbl
\begin{thebibliography}{97}
\providecommand{\natexlab}[1]{#1}
\providecommand{\url}[1]{\texttt{#1}}
\expandafter\ifx\csname urlstyle\endcsname\relax
  \providecommand{\doi}[1]{doi: #1}\else
  \providecommand{\doi}{doi: \begingroup \urlstyle{rm}\Url}\fi

\bibitem[Assran et~al.(2023)Assran, Duval, Misra, Bojanowski, Vincent, Rabbat, LeCun, and Ballas]{ijepa2023}
Mahmoud Assran, Quentin Duval, Ishan Misra, Piotr Bojanowski, Pascal Vincent, Michael Rabbat, Yann LeCun, and Nicolas Ballas.
\newblock Self-supervised learning from images with a joint-embedding predictive architecture.
\newblock \emph{CVPR}, 2023.

\bibitem[Assran et~al.(2025)Assran, Bardes, Fan, Garrido, Howes, Komeili, Muckley, Rizvi, Roberts, Sinha, et~al.]{vjepa2_2025}
Mido Assran, Adrien Bardes, David Fan, Quentin Garrido, Russell Howes, Mojtaba Komeili, Matthew Muckley, Ammar Rizvi, Claire Roberts, Koustuv Sinha, et~al.
\newblock {V-JEPA 2}: Self-supervised video models enable understanding, prediction and planning.
\newblock \emph{arXiv preprint arXiv:2506.09985}, 2025.

\bibitem[Ayad et~al.(2024)Ayad, R{\"o}fer, Heppert, and Valada]{imagine2touch2024}
Abdallah Ayad, Adrian R{\"o}fer, Nick Heppert, and Abhinav Valada.
\newblock Imagine2touch: Predictive tactile sensing for robotic manipulation using efficient low-dimensional signals.
\newblock \emph{arXiv preprint arXiv:2405.01192}, 2024.

\bibitem[Balestriero and LeCun(2025)]{lejepa2025}
Randall Balestriero and Yann LeCun.
\newblock {LeJEPA}: Provable and scalable self-supervised learning without the heuristics.
\newblock \emph{arXiv preprint arXiv:2511.08544}, 2025.

\bibitem[Bardes et~al.(2024)Bardes, Garrido, Ponce, Chen, Rabbat, LeCun, Assran, and Ballas]{vjepa2024}
Adrien Bardes, Quentin Garrido, Jean Ponce, Xinlei Chen, Michael Rabbat, Yann LeCun, Mahmoud Assran, and Nicolas Ballas.
\newblock Revisiting feature prediction for learning visual representations from video.
\newblock \emph{TMLR}, 2024.

\bibitem[Beyer et~al.(2024)Beyer, Steiner, Pinto, Kolesnikov, Wang, Salz, Neumann, Alabdulmohsin, Tschannen, Bugliarello, et~al.]{paligemma2024}
Lucas Beyer, Andreas Steiner, Andr{\'e}~Susano Pinto, Alexander Kolesnikov, Xiao Wang, Daniel Salz, Maxim Neumann, Ibrahim Alabdulmohsin, Michael Tschannen, Emanuele Bugliarello, et~al.
\newblock {PaliGemma}: A versatile 3{B} {VLM} for transfer.
\newblock \emph{arXiv preprint arXiv:2407.07726}, 2024.

\bibitem[Bi et~al.(2026)Bi, Ma, Hao, Shou, and Soh]{vlatouch2025}
Jianxin Bi, Kevin~Yuchen Ma, Ce Hao, Mike~Zheng Shou, and Harold Soh.
\newblock {VLA-Touch}: Enhancing vision-language-action models with dual-level tactile feedback.
\newblock \emph{RA-L}, 2026.

\bibitem[Bjorck et~al.(2025)Bjorck, Casta{\~n}eda, Cherniadev, Da, Ding, Fan, Fang, Fox, Hu, Huang, et~al.]{groot_n1_2025}
Johan Bjorck, Fernando Casta{\~n}eda, Nikita Cherniadev, Xingye Da, Runyu Ding, Linxi Fan, Yu Fang, Dieter Fox, Fengyuan Hu, Spencer Huang, et~al.
\newblock {GR00T N1}: An open foundation model for generalist humanoid robots.
\newblock \emph{arXiv preprint arXiv:2503.14734}, 2025.

\bibitem[Black et~al.(2025)Black, Brown, Driess, Esmail, Equi, Finn, Fusai, Groom, Hausman, Ichter, et~al.]{pi0_2024}
Kevin Black, Noah Brown, Danny Driess, Adnan Esmail, Michael Equi, Chelsea Finn, Niccolo Fusai, Lachy Groom, Karol Hausman, Brian Ichter, et~al.
\newblock {$\pi_0$}: A vision-language-action flow model for general robot control.
\newblock \emph{RSS}, 2025.

\bibitem[Brohan et~al.(2023{\natexlab{a}})Brohan, Brown, Carbajal, Chebotar, Chen, Choromanski, Ding, Driess, Dubey, Finn, et~al.]{rt2_2023}
Anthony Brohan, Noah Brown, Justice Carbajal, Yevgen Chebotar, Xi Chen, Krzysztof Choromanski, Tianli Ding, Danny Driess, Avinava Dubey, Chelsea Finn, et~al.
\newblock {RT-2}: Vision-language-action models transfer web knowledge to robotic control.
\newblock \emph{CoRL}, 2023{\natexlab{a}}.

\bibitem[Brohan et~al.(2023{\natexlab{b}})Brohan, Brown, Carbajal, Chebotar, Dabis, Finn, Gopalakrishnan, Hausman, Herzog, Hsu, et~al.]{rt1}
Anthony Brohan, Noah Brown, Justice Carbajal, Yevgen Chebotar, Joseph Dabis, Chelsea Finn, Keerthana Gopalakrishnan, Karol Hausman, Alexander Herzog, Jasmine Hsu, et~al.
\newblock {RT-1}: Robotics transformer for real-world control at scale.
\newblock \emph{RSS}, 2023{\natexlab{b}}.

\bibitem[Bu et~al.(2025)Bu, Yang, Cai, Gao, Ren, Yao, Luo, and Li]{univla2025}
Qingwen Bu, Yanting Yang, Jisong Cai, Shenyuan Gao, Guanghui Ren, Maoqing Yao, Ping Luo, and Hongyang Li.
\newblock {UniVLA}: Learning to act anywhere with task-centric latent actions.
\newblock \emph{RSS}, 2025.

\bibitem[Cadene et~al.(2026)Cadene, Aliberts, Capuano, Aractingi, Zouitine, Kooijmans, Choghari, Russi, Pascal, Palma, et~al.]{lerobot2024}
Remi Cadene, Simon Aliberts, Francesco Capuano, Michel Aractingi, Adil Zouitine, Pepijn Kooijmans, Jade Choghari, Martino Russi, Caroline Pascal, Steven Palma, et~al.
\newblock {LeRobot}: An open-source library for end-to-end robot learning.
\newblock \emph{ICLR}, 2026.

\bibitem[Cai et~al.(2026{\natexlab{a}})Cai, Cai, Cao, Chen, He, Jiang, Li, Li, Li, Liu, et~al.]{cai2026internvla}
Junhao Cai, Zetao Cai, Jiafei Cao, Yilun Chen, Zeyu He, Lei Jiang, Hang Li, Hengjie Li, Yang Li, Yufei Liu, et~al.
\newblock {InternVLA-A1}: Unifying understanding, generation and action for robotic manipulation.
\newblock \emph{arXiv preprint arXiv:2601.02456}, 2026{\natexlab{a}}.

\bibitem[Cai et~al.(2026{\natexlab{b}})Cai, Guo, He, Jin, Li, Lin, Liu, Liu, Ma, Ma, et~al.]{cai2026xiaomirobotics}
Rui Cai, Jun Guo, Xinze He, Piaopiao Jin, Jie Li, Bingxuan Lin, Futeng Liu, Wei Liu, Fei Ma, Kun Ma, et~al.
\newblock {Xiaomi-Robotics-0}: An open-sourced vision-language-action model with real-time execution.
\newblock \emph{arXiv preprint arXiv:2602.12684}, 2026{\natexlab{b}}.

\bibitem[Cheang et~al.(2025)Cheang, Chen, Cui, Hu, Huang, Kong, Li, Li, Liu, Ma, et~al.]{gr3_2025}
Chilam Cheang, Sijin Chen, Zhongren Cui, Yingdong Hu, Liqun Huang, Tao Kong, Hang Li, Yifeng Li, Yuxiao Liu, Xiao Ma, et~al.
\newblock {GR-3} technical report.
\newblock \emph{arXiv preprint arXiv:2507.15493}, 2025.

\bibitem[Cheang et~al.(2024)Cheang, Chen, Jing, Kong, Li, Li, Liu, Wu, Xu, Yang, et~al.]{gr2}
Chi-Lam Cheang, Guangzeng Chen, Ya Jing, Tao Kong, Hang Li, Yifeng Li, Yuxiao Liu, Hongtao Wu, Jiafeng Xu, Yichu Yang, et~al.
\newblock {GR-2}: A generative video-language-action model with web-scale knowledge for robot manipulation.
\newblock \emph{arXiv preprint arXiv:2410.06158}, 2024.

\bibitem[Chen et~al.(2026{\natexlab{a}})Chen, Wan, Chen, Guo, Xu, Qi, Zhang, Wu, Xu, Li, et~al.]{univtac2026}
Baijun Chen, Weijie Wan, Tianxing Chen, Xianda Guo, Congsheng Xu, Yuanyang Qi, Haojie Zhang, Longyan Wu, Tianling Xu, Zixuan Li, et~al.
\newblock {UniVTAC}: A unified simulation platform for visuo-tactile manipulation data generation, learning, and benchmarking.
\newblock \emph{arXiv preprint arXiv:2602.10093}, 2026{\natexlab{a}}.

\bibitem[Chen et~al.(2026{\natexlab{b}})Chen, Xu, Chen, Hong, Huang, Liu, Mao, Li, Du, and Driggs-Campbell]{policyconsensus2025}
Haonan Chen, Jiaming Xu, Hongyu Chen, Kaiwen Hong, Binghao Huang, Chaoqi Liu, Jiayuan Mao, Yunzhu Li, Yilun Du, and Katherine Driggs-Campbell.
\newblock Multi-modal manipulation via multi-modal policy consensus.
\newblock \emph{ICRA}, 2026{\natexlab{b}}.

\bibitem[Chen et~al.(2026{\natexlab{c}})Chen, Yu, Schwager, Abbeel, Shentu, and Wu]{sarm2026}
Qianzhong Chen, Justin Yu, Mac Schwager, Pieter Abbeel, Yide Shentu, and Philipp Wu.
\newblock {SARM}: Stage-aware reward modeling for long horizon robot manipulation.
\newblock \emph{ICLR}, 2026{\natexlab{c}}.

\bibitem[Chen et~al.(2022)Chen, Sipos, Van~der Merwe, and Fazeli]{vtt2022}
Yizhou Chen, Andrea Sipos, Mark Van~der Merwe, and Nima Fazeli.
\newblock Visuo-tactile transformers for manipulation.
\newblock \emph{CoRL}, 2022.

\bibitem[Cheng et~al.(2026)Cheng, Zhang, Tang, Wang, Zhang, Li, Zhang, and Song]{omnivtla2025}
Zhengxue Cheng, Yiqian Zhang, Anni Tang, Keyu Wang, Wenkang Zhang, Haoyu Li, Hengdi Zhang, and Li Song.
\newblock {OmniVTLA}: Vision-tactile-language-action models with semantic-aligned tactile sensing.
\newblock \emph{RA-L}, 2026.

\bibitem[Chi et~al.(2023)Chi, Xu, Feng, Cousineau, Du, Burchfiel, Tedrake, and Song]{diffusionpolicy2023}
Cheng Chi, Zhenjia Xu, Siyuan Feng, Eric Cousineau, Yilun Du, Benjamin Burchfiel, Russ Tedrake, and Shuran Song.
\newblock Diffusion policy: Visuomotor policy learning via action diffusion.
\newblock \emph{RSS}, 2023.

\bibitem[Chi et~al.(2024)Chi, Xu, Pan, Cousineau, Burchfiel, Feng, Tedrake, and Song]{umi2024}
Cheng Chi, Zhenjia Xu, Chuer Pan, Eric Cousineau, Benjamin Burchfiel, Siyuan Feng, Russ Tedrake, and Shuran Song.
\newblock Universal manipulation interface: In-the-wild robot teaching without in-the-wild robots.
\newblock \emph{RSS}, 2024.

\bibitem[DeepMind(2025)]{gemini35flash}
Google DeepMind.
\newblock Gemini 3.5 flash.
\newblock \url{https://deepmind.google/models/gemini/flash/}, 2025.

\bibitem[Du et~al.(2023)Du, Yang, Dai, Dai, Nachum, Tenenbaum, Schuurmans, and Abbeel]{unipi}
Yilun Du, Mengjiao Yang, Bo Dai, Hanjun Dai, Ofir Nachum, Joshua~B. Tenenbaum, Dale Schuurmans, and Pieter Abbeel.
\newblock Learning universal policies via text-guided video generation.
\newblock \emph{NeurIPS}, 2023.

\bibitem[Fei et~al.(2026)Fei, Wang, Shi, Dai, Cai, Qian, Ji, He, Zhang, Fei, Fu, Gong, and Qiu]{liberoplus2025}
Senyu Fei, Siyin Wang, Junhao Shi, Zihao Dai, Jikun Cai, Pengfang Qian, Li Ji, Xinzhe He, Shiduo Zhang, Zhaoye Fei, Jinlan Fu, Jingjing Gong, and Xipeng Qiu.
\newblock {LIBERO-Plus}: In-depth robustness analysis of vision-language-action models.
\newblock \emph{CVPR}, 2026.

\bibitem[Feng et~al.(2025)Feng, Hu, Xia, Gao, Shen, Sun, Fang, and Hu]{anytouch2025}
Ruoxuan Feng, Jiangyu Hu, Wenke Xia, Tianci Gao, Ao Shen, Yuhao Sun, Bin Fang, and Di Hu.
\newblock {AnyTouch}: Learning unified static-dynamic representation across multiple visuo-tactile sensors.
\newblock \emph{ICLR}, 2025.

\bibitem[George et~al.(2025)George, Gano, Katragadda, and Barati~Farimani]{vital2024}
Abraham George, Selam Gano, Pranav Katragadda, and Amir Barati~Farimani.
\newblock {VITaL} pretraining: Visuo-tactile pretraining for tactile and non-tactile manipulation policies.
\newblock \emph{ICRA}, 2025.

\bibitem[Gubernatorov et~al.(2026)Gubernatorov, Sannikov, Mikhalchuk, Kuznetsov, Artemov, Ouwatobi, Fernando, Asanov, Guo, and Tsetserukou]{hapticvla2026}
Konstantin Gubernatorov, Mikhail Sannikov, Ilya Mikhalchuk, Egor Kuznetsov, Makar Artemov, Ogunwoye~Faith Ouwatobi, Marcelino Fernando, Artem Asanov, Ziang Guo, and Dzmitry Tsetserukou.
\newblock {HapticVLA}: Contact-rich manipulation via vision-language-action model without inference-time tactile sensing.
\newblock \emph{arXiv preprint arXiv:2603.15257}, 2026.

\bibitem[Hao et~al.(2026)Hao, Zhang, Li, Cao, Hao, Cui, and Wang]{tla2025}
Peng Hao, Chaofan Zhang, Dingzhe Li, Xiaoge Cao, Xiaoshuai Hao, Shaowei Cui, and Shuo Wang.
\newblock {TLA}: Tactile-language-action model for contact-rich manipulation.
\newblock \emph{Robot Learning}, 2026.

\bibitem[Hao et~al.(2025)Hao, Zhou, Huang, Hou, Tang, Zhang, Li, Lu, Ren, Meng, et~al.]{mimoembodied2025}
Xiaoshuai Hao, Lei Zhou, Zhijian Huang, Zhiwen Hou, Yingbo Tang, Lingfeng Zhang, Guang Li, Zheng Lu, Shuhuai Ren, Xianhui Meng, et~al.
\newblock {MiMo-Embodied}: X-embodied foundation model technical report.
\newblock \emph{arXiv preprint arXiv:2511.16518}, 2025.

\bibitem[Higuera et~al.(2024)Higuera, Sharma, Bodduluri, Fan, Lancaster, Kalakrishnan, Kaess, Boots, Lambeta, Wu, et~al.]{sparsh2024}
Carolina Higuera, Akash Sharma, Chaithanya~Krishna Bodduluri, Taosha Fan, Patrick Lancaster, Mrinal Kalakrishnan, Michael Kaess, Byron Boots, Mike Lambeta, Tingfan Wu, et~al.
\newblock Sparsh: Self-supervised touch representations for vision-based tactile sensing.
\newblock \emph{CoRL}, 2024.

\bibitem[Higuera et~al.(2026)Higuera, Arnaud, Boots, Mukadam, Hogan, and Meier]{vtwm2026}
Carolina Higuera, Sergio Arnaud, Byron Boots, Mustafa Mukadam, Francois~Robert Hogan, and Franziska Meier.
\newblock Visuo-tactile world models.
\newblock \emph{arXiv preprint arXiv:2602.06001}, 2026.

\bibitem[Huang et~al.(2024)Huang, Wang, Yang, Luo, and Li]{vitac3d2024}
Binghao Huang, Yixuan Wang, Xinyi Yang, Yiyue Luo, and Yunzhu Li.
\newblock {3D-ViTac}: Learning fine-grained manipulation with visuo-tactile sensing.
\newblock \emph{CoRL}, 2024.

\bibitem[Huang et~al.(2025)Huang, Wang, Lin, Hu, Wen, and Gao]{tactilevla2025}
Jialei Huang, Shuo Wang, Fanqi Lin, Yihang Hu, Chuan Wen, and Yang Gao.
\newblock {Tactile-VLA}: Unlocking vision-language-action model's physical knowledge for tactile generalization.
\newblock \emph{arXiv preprint arXiv:2507.09160}, 2025.

\bibitem[Kim et~al.(2024)Kim, Pertsch, Karamcheti, Xiao, Balakrishna, Nair, Rafailov, Foster, Lam, Sanketi, et~al.]{openvla2024}
Moo~Jin Kim, Karl Pertsch, Siddharth Karamcheti, Ted Xiao, Ashwin Balakrishna, Suraj Nair, Rafael Rafailov, Ethan Foster, Grace Lam, Pannag Sanketi, et~al.
\newblock {OpenVLA}: An open-source vision-language-action model.
\newblock \emph{CoRL}, 2024.

\bibitem[Kim et~al.(2026)Kim, Choi, Baek, and Renders]{progvla2026}
Seungsu Kim, Jinyoung Choi, Seungmin Baek, and Jean-Michel Renders.
\newblock {ProgVLA}: Progress-aware robot manipulation skill learning.
\newblock \emph{arXiv preprint arXiv:2605.28231}, 2026.

\bibitem[Lambeta et~al.(2020)Lambeta, Chou, Tian, Yang, Maloon, Most, Stroud, Santos, Byagowi, Kammerer, et~al.]{digit2020}
Mike Lambeta, Po-Wei Chou, Stephen Tian, Brian Yang, Benjamin Maloon, Victoria~Rose Most, Dave Stroud, Raymond Santos, Ahmad Byagowi, Gregg Kammerer, et~al.
\newblock {DIGIT}: A novel design for a low-cost compact high-resolution tactile sensor with application to in-hand manipulation.
\newblock \emph{RA-L}, 2020.

\bibitem[Li et~al.(2026{\natexlab{a}})Li, Yang, Chen, Chen, Yang, Tian, Wang, Wang, Lin, Zhao, et~al.]{cronusvla}
Hao Li, Shuai Yang, Yilun Chen, Xinyi Chen, Xiaoda Yang, Yang Tian, Hanqing Wang, Tai Wang, Dahua Lin, Feng Zhao, et~al.
\newblock {CronusVLA}: Towards efficient and robust manipulation via multi-frame vision-language-action modeling.
\newblock \emph{AAAI}, 2026{\natexlab{a}}.

\bibitem[Li et~al.(2026{\natexlab{b}})Li, Guo, Li, Qian, Lai, Wang, Yan, Cao, Chen, Qu, et~al.]{li2026xiaomiu0}
Xinghang Li, Jun Guo, Qiwei Li, Long Qian, Hang Lai, Yueze Wang, Hongyu Yan, Jiahang Cao, Xi Chen, Jingen Qu, et~al.
\newblock {Xiaomi-Robotics-U0}: Unified embodied synthesis with world foundation model.
\newblock \emph{arXiv preprint arXiv:2607.11643}, 2026{\natexlab{b}}.

\bibitem[Li et~al.(2025)Li, Ma, Xu, Cui, Cui, Han, Huang, Kong, Liu, Niu, et~al.]{grrl2025}
Yunfei Li, Xiao Ma, Jiafeng Xu, Yu Cui, Zhongren Cui, Zhigang Han, Liqun Huang, Tao Kong, Yuxiao Liu, Hao Niu, et~al.
\newblock {GR-RL}: Going dexterous and precise for long-horizon robotic manipulation.
\newblock \emph{arXiv preprint arXiv:2512.01801}, 2025.

\bibitem[Lipman et~al.(2023)Lipman, Chen, Ben-Hamu, Nickel, and Le]{flowmatching2022}
Yaron Lipman, Ricky T.~Q. Chen, Heli Ben-Hamu, Maximilian Nickel, and Matt Le.
\newblock Flow matching for generative modeling.
\newblock \emph{ICLR}, 2023.

\bibitem[Liu et~al.(2025)Liu, Li, Qin, Xu, Abbeel, and Chen]{vitamin2025}
Fangchen Liu, Chuanyu Li, Yihua Qin, Jing Xu, Pieter Abbeel, and Rui Chen.
\newblock {ViTaMIn}: Learning contact-rich tasks through robot-free visuo-tactile manipulation interface.
\newblock \emph{arXiv preprint arXiv:2504.06156}, 2025.

\bibitem[Lou et~al.(2026)Lou, Ye, Fu, Cen, Chi, Lyu, Jia, Han, Lu, and Zhang]{dreamtac2026}
Yunfan Lou, Yifan Ye, Yankai Fu, Jun Cen, Xiaowei Chi, Yaoxu Lyu, Peidong Jia, Sirui Han, Zhihe Lu, and Shanghang Zhang.
\newblock {Dream-Tac}: A unified tactile world action model for contact-rich robot manipulation.
\newblock \emph{arXiv preprint arXiv:2606.08737}, 2026.

\bibitem[Lu et~al.(2025)Lu, Guo, Zhang, Zhou, Jiang, Gao, Tang, and Wang]{vlarl2025}
Guanxing Lu, Wenkai Guo, Chubin Zhang, Yuheng Zhou, Haonan Jiang, Zifeng Gao, Yansong Tang, and Ziwei Wang.
\newblock {VLA-RL}: Towards masterful and general robotic manipulation with scalable reinforcement learning.
\newblock \emph{arXiv preprint arXiv:2505.18719}, 2025.

\bibitem[Ma et~al.(2026)Ma, Cai, Xu, Li, Yang, Tian, Cao, Zhu, Qiu, {Zhaxizhuoma}, et~al.]{internvla_a15}
Haoxiang Ma, Junhao Cai, Xiaoxu Xu, Hao Li, Yuyin Yang, Yang Tian, Jiafei Cao, Hongrui Zhu, Zherui Qiu, {Zhaxizhuoma}, et~al.
\newblock {InternVLA-A1.5}: Unifying understanding, latent foresight, and action for compositional generalization.
\newblock \emph{arXiv preprint arXiv:2607.04988}, 2026.

\bibitem[Ma et~al.(2025)Ma, Hejna, Wahid, Fu, Shah, Liang, Xu, Kirmani, Xu, Driess, et~al.]{gvl2025}
Yecheng~Jason Ma, Joey Hejna, Ayzaan Wahid, Chuyuan Fu, Dhruv Shah, Jacky Liang, Zhuo Xu, Sean Kirmani, Peng Xu, Danny Driess, et~al.
\newblock Vision language models are in-context value learners.
\newblock \emph{ICLR}, 2025.

\bibitem[Maes et~al.(2026)Maes, Le~Lidec, Scieur, LeCun, and Balestriero]{leworldmodel2026}
Lucas Maes, Quentin Le~Lidec, Damien Scieur, Yann LeCun, and Randall Balestriero.
\newblock {LeWorldModel}: Stable end-to-end joint-embedding predictive architecture from pixels.
\newblock \emph{arXiv preprint arXiv:2603.19312}, 2026.

\bibitem[Mao et~al.(2026)Mao, Yu, Mao, Li, Hu, Lan, Zhu, and Chen]{arm2026}
Yiming Mao, Zixi Yu, Weixin Mao, Yinhao Li, Qirui Hu, Zihan Lan, Minzhao Zhu, and Hua Chen.
\newblock {ARM}: Advantage reward modeling for long-horizon manipulation.
\newblock \emph{arXiv preprint arXiv:2604.03037}, 2026.

\bibitem[{NeoteAI Team} and {Fudan TEAI Team}(2026)]{neoteaidata2026}
{NeoteAI Team} and {Fudan TEAI Team}.
\newblock {$\mathcal{N}_0$}-foundation: Towards the age of tactile intelligence.
\newblock \url{https://research.neoteai.com/n0-foundation/}, 2026.

\bibitem[Niu et~al.(2026)Niu, Liu, Wang, Shao, Yin, Pai, Sharma, Saravalle, Zheng, Wang, et~al.]{trex2026}
Dantong Niu, Zhuoyang Liu, Zekai Wang, Boning Shao, Zhao-Heng Yin, Anirudh Pai, Yuvan Sharma, Stefano Saravalle, Ruijie Zheng, Jing Wang, et~al.
\newblock {T-Rex}: Tactile-reactive dexterous manipulation.
\newblock \emph{arXiv preprint arXiv:2606.17055}, 2026.

\bibitem[{Octo Model Team} et~al.(2024){Octo Model Team}, Ghosh, Walke, Pertsch, Black, Mees, Dasari, Hejna, Kreiman, Xu, Luo, et~al.]{octo2024}
{Octo Model Team}, Dibya Ghosh, Homer Walke, Karl Pertsch, Kevin Black, Oier Mees, Sudeep Dasari, Joey Hejna, Tobias Kreiman, Charles Xu, Jianlan Luo, et~al.
\newblock Octo: An open-source generalist robot policy.
\newblock \emph{RSS}, 2024.

\bibitem[{Open X-Embodiment Collaboration} et~al.(2024){Open X-Embodiment Collaboration}, O'Neill, Rehman, Gupta, Maddukuri, Gupta, Padalkar, Lee, Pooley, Gupta, Mandlekar, et~al.]{openx2023}
{Open X-Embodiment Collaboration}, Abby O'Neill, Abdul Rehman, Abhinav Gupta, Abhiram Maddukuri, Abhishek Gupta, Abhishek Padalkar, Abraham Lee, Acorn Pooley, Agrim Gupta, Ajay Mandlekar, et~al.
\newblock Open {X}-embodiment: Robotic learning datasets and {RT-X} models.
\newblock \emph{ICRA}, 2024.

\bibitem[Oquab et~al.(2024)Oquab, Darcet, Moutakanni, Vo, Szafraniec, Khalidov, Fernandez, Haziza, Massa, El-Nouby, et~al.]{dinov2_2023}
Maxime Oquab, Timoth{\'e}e Darcet, Th{\'e}o Moutakanni, Huy Vo, Marc Szafraniec, Vasil Khalidov, Pierre Fernandez, Daniel Haziza, Francisco Massa, Alaaeldin El-Nouby, et~al.
\newblock {DINOv2}: Learning robust visual features without supervision.
\newblock \emph{TMLR}, 2024.

\bibitem[Pan et~al.(2026)Pan, Feng, Zhang, Li, Song, Qu, Wang, Li, Xiong, Chen, et~al.]{sop2026}
Mingjie Pan, Siyuan Feng, Qinglin Zhang, Xinchen Li, Jianheng Song, Chendi Qu, Yi Wang, Chuankang Li, Ziyu Xiong, Zhi Chen, et~al.
\newblock {SOP}: A scalable online post-training system for vision-language-action models.
\newblock \emph{arXiv preprint arXiv:2601.03044}, 2026.

\bibitem[{Physical Intelligence} et~al.(2025{\natexlab{a}}){Physical Intelligence}, Amin, Aniceto, Balakrishna, Black, Conley, Connors, Darpinian, Dhabalia, DiCarlo, Driess, et~al.]{pistar06_2025}
{Physical Intelligence}, Ali Amin, Raichelle Aniceto, Ashwin Balakrishna, Kevin Black, Ken Conley, Grace Connors, James Darpinian, Karan Dhabalia, Jared DiCarlo, Danny Driess, et~al.
\newblock {$\pi^{*}_{0.6}$}: a {VLA} that learns from experience.
\newblock \emph{arXiv preprint arXiv:2511.14759}, 2025{\natexlab{a}}.

\bibitem[{Physical Intelligence} et~al.(2025{\natexlab{b}}){Physical Intelligence}, Black, Brown, Darpinian, Dhabalia, Driess, Esmail, Equi, Finn, Fusai, Galliker, et~al.]{pi05_2025}
{Physical Intelligence}, Kevin Black, Noah Brown, James Darpinian, Karan Dhabalia, Danny Driess, Adnan Esmail, Michael Equi, Chelsea Finn, Niccolo Fusai, Manuel~Y. Galliker, et~al.
\newblock {$\pi_{0.5}$}: a vision-language-action model with open-world generalization.
\newblock \emph{CoRL}, 2025{\natexlab{b}}.

\bibitem[Shou et~al.(2021)Shou, Lei, Wang, Ghadiyaram, and Feiszli]{shou2021gebd}
Mike~Zheng Shou, Stan~Weixian Lei, Weiyao Wang, Deepti Ghadiyaram, and Matt Feiszli.
\newblock Generic event boundary detection: A benchmark for event segmentation.
\newblock \emph{ICCV}, 2021.

\bibitem[Shukor et~al.(2025)Shukor, Aubakirova, Capuano, Kooijmans, Palma, Zouitine, Aractingi, Pascal, Russi, Marafioti, et~al.]{smolvla2025}
Mustafa Shukor, Dana Aubakirova, Francesco Capuano, Pepijn Kooijmans, Steven Palma, Adil Zouitine, Michel Aractingi, Caroline Pascal, Martino Russi, Andres Marafioti, et~al.
\newblock {SmolVLA}: A vision-language-action model for affordable and efficient robotics.
\newblock \emph{arXiv preprint arXiv:2506.01844}, 2025.

\bibitem[{StarVLA Community}(2026)]{starvla2026}
{StarVLA Community}.
\newblock {StarVLA}: A lego-like codebase for vision-language-action model developing.
\newblock \emph{arXiv preprint arXiv:2604.05014}, 2026.

\bibitem[Su et~al.(2026)Su, Chen, Shi, Liu, Zhang, Huang, Zhong, Zhu, Liu, and Liu]{WoG}
Yue Su, Sijin Chen, Haixin Shi, Mingyu Liu, Zhengshen Zhang, Ningyuan Huang, Weiheng Zhong, Zhengbang Zhu, Yuxiao Liu, and Xihui Liu.
\newblock World guidance: World modeling in condition space for action generation.
\newblock \emph{arXiv preprint arXiv:2602.22010}, 2026.

\bibitem[Suresh et~al.(2024)Suresh, Qi, Wu, Fan, Pineda, Lambeta, Malik, Kalakrishnan, Calandra, Kaess, et~al.]{neuralfeels2024}
Sudharshan Suresh, Haozhi Qi, Tingfan Wu, Taosha Fan, Luis Pineda, Mike Lambeta, Jitendra Malik, Mrinal Kalakrishnan, Roberto Calandra, Michael Kaess, et~al.
\newblock {NeuralFeels} with neural fields: Visuotactile perception for in-hand manipulation.
\newblock \emph{Sci. Robot.}, 2024.

\bibitem[Tan et~al.(2026)Tan, Chen, Xu, Wang, Ji, Chi, Lyu, Zhao, Chen, Co, et~al.]{robodopamine2025}
Huajie Tan, Sixiang Chen, Yijie Xu, Zixiao Wang, Yuheng Ji, Cheng Chi, Yaoxu Lyu, Zhongxia Zhao, Xiansheng Chen, Peterson Co, et~al.
\newblock {Robo-Dopamine}: General process reward modeling for high-precision robotic manipulation.
\newblock \emph{CVPR}, 2026.

\bibitem[Tian et~al.(2026)Tian, Zheng, Zheng, Gu, Zang, Qin, Li, Li, Ding, and Zhao]{vtwam2026}
Shuai Tian, Yupeng Zheng, Yuhang Zheng, Songen Gu, Yujie Zang, Yuxing Qin, Weize Li, Haoran Li, Wenchao Ding, and Dongbin Zhao.
\newblock {VT-WAM}: Visual-tactile world action model for contact-rich manipulation.
\newblock \emph{arXiv preprint arXiv:2607.02503}, 2026.

\bibitem[van~den Oord et~al.(2018)van~den Oord, Li, and Vinyals]{infonce2018}
A{\"a}ron van~den Oord, Yazhe Li, and Oriol Vinyals.
\newblock Representation learning with contrastive predictive coding.
\newblock \emph{arXiv preprint arXiv:1807.03748}, 2018.

\bibitem[Wang et~al.(2026{\natexlab{a}})Wang, Li, Guan, Ye, Xie, Liu, Chen, Liang, Zhang, Hu, et~al.]{qwenvla2026}
Qiuyue Wang, Mingsheng Li, Jian Guan, Jinhui Ye, Sicheng Xie, Yitao Liu, Junhao Chen, Zhixuan Liang, Jie Zhang, Xintong Hu, et~al.
\newblock {Qwen-VLA}: Unifying vision-language-action modeling across tasks, environments, and robot embodiments.
\newblock \emph{arXiv preprint arXiv:2605.30280}, 2026{\natexlab{a}}.

\bibitem[Wang et~al.(2026{\natexlab{b}})Wang, Li, Xie, Yang, Nie, Cai, Zhang, Qu, Wu, Song, et~al.]{lwd2026}
Yi Wang, Xinchen Li, Pengwei Xie, Pu Yang, Buqing Nie, Yunuo Cai, Qinglin Zhang, Chendi Qu, Jeffrey Wu, Jianheng Song, et~al.
\newblock Learning while deploying: Fleet-scale reinforcement learning for generalist robot policies.
\newblock \emph{arXiv preprint arXiv:2605.00416}, 2026{\natexlab{b}}.

\bibitem[Wu et~al.(2024)Wu, Jing, Cheang, Chen, Xu, Li, Liu, Li, and Kong]{gr1}
Hongtao Wu, Ya Jing, Chilam Cheang, Guangzeng Chen, Jiafeng Xu, Xinghang Li, Minghuan Liu, Hang Li, and Tao Kong.
\newblock Unleashing large-scale video generative pre-training for visual robot manipulation.
\newblock \emph{ICLR}, 2024.

\bibitem[Wu et~al.(2026{\natexlab{a}})Wu, Yu, Ren, Chen, Jiang, Huang, Gu, and Li]{freetacman2025}
Longyan Wu, Checheng Yu, Jieji Ren, Li Chen, Yufei Jiang, Ran Huang, Guoying Gu, and Hongyang Li.
\newblock {FreeTacMan}: Robot-free visuo-tactile data collection system for contact-rich manipulation.
\newblock \emph{ICRA}, 2026{\natexlab{a}}.

\bibitem[Wu et~al.(2026{\natexlab{b}})Wu, You, Zhu, Liu, Zhang, Liu, Wang, Li, Li, and Zhao]{tactilewam2026}
Siyu Wu, Linjing You, Junjie Zhu, Yaozu Liu, Changhao Zhang, Jian Liu, Weiqiang Wang, Qi Li, Jituo Li, and Hengshuang Zhao.
\newblock {Tactile-WAM}: Touch-aware world action model with tactile asymmetric attention.
\newblock \emph{arXiv preprint arXiv:2606.26663}, 2026{\natexlab{b}}.

\bibitem[Wu et~al.(2026{\natexlab{c}})Wu, Lu, Wang, Yang, Liu, Wang, Zhu, Sun, Wang, Ma, et~al.]{pragmaticvla2026}
Wei Wu, Fan Lu, Yunnan Wang, Shuai Yang, Shi Liu, Fangjing Wang, Qian Zhu, He Sun, Yong Wang, Shuailei Ma, et~al.
\newblock A pragmatic {VLA} foundation model.
\newblock \emph{arXiv preprint arXiv:2601.18692}, 2026{\natexlab{c}}.

\bibitem[Xia et~al.(2026)Xia, Ren, Yu, Zhang, Li, Zhang, Zhao, Gao, Fu, Tang, et~al.]{robovalue2026}
Wenke Xia, Pei Ren, Wenbo Yu, Yizhuo Zhang, Jifan Li, Yixue Zhang, Yinuo Zhao, Qingyang Gao, Jianlong Fu, Jian Tang, et~al.
\newblock {Robo-ValueRL}: Reliable value estimation for offline-to-online reinforcement learning.
\newblock \emph{arXiv preprint arXiv:2607.09866}, 2026.

\bibitem[{Xiaomi Robotics Team} et~al.(2026){Xiaomi Robotics Team}, Guo, Jin, Li, Li, Li, Liu, Peng, Qin, Su, Sun, et~al.]{xiaomi2026robotics1}
{Xiaomi Robotics Team}, Jun Guo, Piaopiao Jin, Jason Li, Peiyan Li, Yingyan Li, Futeng Liu, Wanli Peng, Optimus Qin, Yifei Su, Nan Sun, et~al.
\newblock {Xiaomi-Robotics-1}: Scaling vision-language-action models with over 100{K} hours of real-world trajectories.
\newblock \emph{arXiv preprint arXiv:2607.15330}, 2026.

\bibitem[Xu et~al.(2026)Xu, Zhong, Liu, Wang, Luo, Zhou, Zhang, Ye, Wu, and Jiang]{seeingtouch2026}
Shengqi Xu, Guojin Zhong, Yang Liu, Fanjie Wang, Hu Luo, Hanyu Zhou, Weiyao Zhang, Ziyi Ye, Zuxuan Wu, and Yu-Gang Jiang.
\newblock Seeing touch from motion: A unified modality-aware visuo-tactile policy with tactile motion correlation.
\newblock \emph{ECCV}, 2026.

\bibitem[Xu et~al.(2025)Xu, Wei, An, Zhang, and Li]{exumi2025}
Yue Xu, Litao Wei, Pengyu An, Qingyu Zhang, and Yong-Lu Li.
\newblock {exUMI}: Extensible robot teaching system with action-aware task-agnostic tactile representation.
\newblock \emph{CoRL}, 2025.

\bibitem[Xue et~al.(2025)Xue, Ren, Chen, Zhang, Fang, Gu, Xu, and Lu]{rdp2025}
Han Xue, Jieji Ren, Wendi Chen, Gu Zhang, Yuan Fang, Guoying Gu, Huazhe Xu, and Cewu Lu.
\newblock Reactive diffusion policy: Slow-fast visual-tactile policy learning for contact-rich manipulation.
\newblock \emph{RSS}, 2025.

\bibitem[Yan et~al.(2026)Yan, Li, Yang, and Mu]{progressvla2026}
Hongyu Yan, Qiwei Li, Jiaolong Yang, and Yadong Mu.
\newblock {ProgressVLA}: Progress-guided diffusion policy for vision-language robotic manipulation.
\newblock \emph{arXiv preprint arXiv:2603.27670}, 2026.

\bibitem[Yang et~al.(2022)Yang, Ma, Zhang, Zhu, Yuan, and Owens]{touchandgo2022}
Fengyu Yang, Chenyang Ma, Jiacheng Zhang, Jing Zhu, Wenzhen Yuan, and Andrew Owens.
\newblock Touch and go: Learning from human-collected vision and touch.
\newblock \emph{NeurIPS}, 2022.

\bibitem[Yang et~al.(2024)Yang, Feng, Chen, Park, Wang, Dou, Zeng, Chen, Gangopadhyay, Owens, et~al.]{unitouch2024}
Fengyu Yang, Chao Feng, Ziyang Chen, Hyoungseob Park, Daniel Wang, Yiming Dou, Ziyao Zeng, Xien Chen, Rit Gangopadhyay, Andrew Owens, et~al.
\newblock Binding touch to everything: Learning unified multimodal tactile representations.
\newblock \emph{CVPR}, 2024.

\bibitem[Ye et~al.(2026{\natexlab{a}})Ye, Wang, Ni, Huang, Zhao, Li, Li, Li, Lv, Liu, et~al.]{ye2026gigaworld}
Angen Ye, Boyuan Wang, Chaojun Ni, Guan Huang, Guosheng Zhao, Hao Li, Hengtao Li, Jie Li, Jindi Lv, Jingyu Liu, et~al.
\newblock {GigaWorld-Policy}: An efficient action-centered world--action model.
\newblock \emph{arXiv preprint arXiv:2603.17240}, 2026{\natexlab{a}}.

\bibitem[Ye et~al.(2026{\natexlab{b}})Ye, Gao, Yang, Zheng, Wang, Chen, Chen, Chen, Liu, and Jia]{ye2026starvla}
Jinhui Ye, Ning Gao, Senqiao Yang, Jinliang Zheng, Zixuan Wang, Yuxin Chen, Pengguang Chen, Yilun Chen, Shu Liu, and Jiaya Jia.
\newblock {StarVLA-$\alpha$}: Reducing complexity in vision-language-action systems.
\newblock \emph{arXiv preprint arXiv:2604.11757}, 2026{\natexlab{b}}.

\bibitem[Yu et~al.(2026)Yu, Sima, Jiang, Zhang, Mai, Li, Wang, Chen, Wu, Chen, et~al.]{chi0_2026}
Checheng Yu, Chonghao Sima, Gangcheng Jiang, Hai Zhang, Haoguang Mai, Hongyang Li, Huijie Wang, Jin Chen, Kaiyang Wu, Li Chen, et~al.
\newblock {$\chi_0$}: Resource-aware robust manipulation via taming distributional inconsistencies.
\newblock \emph{arXiv preprint arXiv:2602.09021}, 2026.

\bibitem[Yu et~al.(2025)Yu, Liu, Yu, Ren, Hao, Ding, Huang, Huang, Song, Cai, et~al.]{forcevla}
Jiawen Yu, Hairuo Liu, Qiaojun Yu, Jieji Ren, Ce Hao, Haitong Ding, Guangyu Huang, Guofan Huang, Yan Song, Panpan Cai, et~al.
\newblock {ForceVLA}: Enhancing {VLA} models with a force-aware {MoE} for contact-rich manipulation.
\newblock \emph{NeurIPS}, 2025.

\bibitem[Yuan et~al.(2026)Yuan, Zhang, Zhou, Chen, Wang, Liu, Niu, Wang, Zhang, Zhang, et~al.]{ftp1_2026}
Chengbo Yuan, Zicheng Zhang, Mingjie Zhou, Wendi Chen, Yi Wang, Zhuoyang Liu, Dantong Niu, Shuo Wang, Hui Zhang, Wenkang Zhang, et~al.
\newblock {FTP-1}: A generalist foundation tactile policy across tactile sensors for contact-rich manipulation.
\newblock \emph{arXiv preprint arXiv:2606.13102}, 2026.

\bibitem[Yuan et~al.(2017)Yuan, Dong, and Adelson]{gelsight2017}
Wenzhen Yuan, Siyuan Dong, and Edward~H. Adelson.
\newblock {GelSight}: High-resolution robot tactile sensors for estimating geometry and force.
\newblock \emph{Sensors}, 2017.

\bibitem[Zang et~al.(2026)Zang, Zheng, Nie, Zheng, Tian, Gu, Gao, Wang, Yan, and Ding]{tacforesight}
Yujie Zang, Yuhang Zheng, Xian Nie, Yupeng Zheng, Shuai Tian, Songen Gu, Chen Gao, Zining Wang, Shuicheng Yan, and Wenchao Ding.
\newblock {TacForeSight}: Force-guided tactile world model for contact-rich manipulation.
\newblock \emph{arXiv preprint arXiv:2606.11184}, 2026.

\bibitem[Zhang et~al.(2026{\natexlab{a}})Zhang, Cai, Xi, Yuan, Luo, Zhang, Zheng, Xu, and Lu]{ttp2026}
Chi Zhang, Penglin Cai, Ziheng Xi, Haoqi Yuan, Hao Luo, Wanpeng Zhang, Sipeng Zheng, Chaoyi Xu, and Zongqing Lu.
\newblock Human-centric transferable tactile pre-training for dexterous robotic manipulation.
\newblock \emph{arXiv preprint arXiv:2607.01067}, 2026{\natexlab{a}}.

\bibitem[Zhang et~al.(2025{\natexlab{a}})Zhang, Xu, Liu, Yu, Li, Gao, Fei, Yin, Wu, Jiang, and Qiu]{vlabench2024}
Shiduo Zhang, Zhe Xu, Peiju Liu, Xiaopeng Yu, Yuan Li, Qinghui Gao, Zhaoye Fei, Zhangyue Yin, Zuxuan Wu, Yu-Gang Jiang, and Xipeng Qiu.
\newblock {VLABench}: A large-scale benchmark for language-conditioned robotics manipulation with long-horizon reasoning tasks.
\newblock \emph{ICCV}, 2025{\natexlab{a}}.

\bibitem[Zhang et~al.(2025{\natexlab{b}})Zhang, Xu, Yang, Yue, Lin, Gao, Wang, and Zhao]{tavla}
Zongzheng Zhang, Haobo Xu, Zhuo Yang, Chenghao Yue, Zehao Lin, Huan-ang Gao, Ziwei Wang, and Hao Zhao.
\newblock {TA-VLA}: Elucidating the design space of torque-aware vision-language-action models.
\newblock \emph{CoRL}, 2025{\natexlab{b}}.

\bibitem[Zhang et~al.(2026{\natexlab{b}})Zhang, Ma, Yang, Wen, Zhang, Li, Qin, Liu, Zhao, Kang, et~al.]{touchguide2026}
Zhemeng Zhang, Jiahua Ma, Xincheng Yang, Xin Wen, Yuzhi Zhang, Boyan Li, Yiran Qin, Jin Liu, Can Zhao, Li Kang, et~al.
\newblock {TouchGuide}: Inference-time steering of visuomotor policies via touch guidance.
\newblock \emph{RSS}, 2026{\natexlab{b}}.

\bibitem[Zhang et~al.(2026{\natexlab{c}})Zhang, Zhou, Zhang, Desai, Amosa, Soleymanzadeh, Lei, Zheng, and She]{contactworld}
Zhiyuan Zhang, Pokuang Zhou, Kaidi Zhang, Adeesh Desai, Temitope Amosa, Davood Soleymanzadeh, Jiuzhou Lei, Minghui Zheng, and Yu She.
\newblock {ContactWorld}: What matters in vision-tactile world models for contact-rich manipulation.
\newblock \emph{arXiv preprint arXiv:2606.13877}, 2026{\natexlab{c}}.

\bibitem[Zhao et~al.(2023)Zhao, Kumar, Levine, and Finn]{act2023}
Tony~Z. Zhao, Vikash Kumar, Sergey Levine, and Chelsea Finn.
\newblock Learning fine-grained bimanual manipulation with low-cost hardware.
\newblock \emph{RSS}, 2023.

\bibitem[Zheng et~al.(2026)Zheng, Gu, Li, Zheng, Zang, Tian, Li, Hao, Gao, Liu, et~al.]{omnivta2026}
Yuhang Zheng, Songen Gu, Weize Li, Yupeng Zheng, Yujie Zang, Shuai Tian, Xiang Li, Ce Hao, Chen Gao, Si Liu, et~al.
\newblock {OmniVTA}: Visuo-tactile world modeling for contact-rich robotic manipulation.
\newblock \emph{arXiv preprint arXiv:2603.19201}, 2026.

\bibitem[Zhou et~al.(2019)Zhou, Barnes, Lu, Yang, and Li]{zhou2019rot6d}
Yi Zhou, Connelly Barnes, Jingwan Lu, Jimei Yang, and Hao Li.
\newblock On the continuity of rotation representations in neural networks.
\newblock \emph{CVPR}, 2019.

\bibitem[Zhu et~al.(2025{\natexlab{a}})Zhu, Yu, Feng, Burchfiel, Shah, and Gupta]{uwm}
Chuning Zhu, Raymond Yu, Siyuan Feng, Benjamin Burchfiel, Paarth Shah, and Abhishek Gupta.
\newblock Unified world models: Coupling video and action diffusion for pretraining on large robotic datasets.
\newblock \emph{RSS}, 2025{\natexlab{a}}.

\bibitem[Zhu et~al.(2025{\natexlab{b}})Zhu, Huang, and Li]{touchinthewild2025}
Xinyue Zhu, Binghao Huang, and Yunzhu Li.
\newblock Touch in the wild: Learning fine-grained manipulation with a portable visuo-tactile gripper.
\newblock \emph{NeurIPS}, 2025{\natexlab{b}}.

\end{thebibliography}
